\documentclass[lettersize,journal]{IEEEtran}
\usepackage{amsmath,amsfonts}
\usepackage{algorithmic}
\usepackage{algorithm}
\usepackage{array}
\usepackage[caption=false,font=normalsize,labelfont=sf,textfont=sf]{subfig}
\usepackage{textcomp}
\usepackage{stfloats}
\usepackage{url}
\usepackage{verbatim}
\usepackage{xcolor}

\usepackage{graphicx}
\usepackage{makecell}
\usepackage{multirow}
\usepackage{eqexpl}
\usepackage{booktabs}
\eqexplSetDelim{:}
\usepackage{cite}
\newcommand\tab[1][1cm]{\hspace*{#1}}

\begin{document}
\title{Run-time Introspection of 2D Object Detection in Automated Driving Systems Using Learning Representations}
\author{Hakan Yekta Yatbaz,~\IEEEmembership{Student Member,~IEEE},~Mehrdad Dianati,~\IEEEmembership{Senior Member,~IEEE}, Konstantinos Koufos and Roger Woodman
\thanks{This research has been conducted as part of the EVENTS project, which is funded by the European Union, under grant agreement No 101069614. Views and opinions expressed are however those
of the author(s) only and do not necessarily reflect those of the European Union or European Commission. Neither the European Union nor the granting authority can be held responsible for them. This research has been also supported by the Centre for Doctoral Training (CDT) to Advance the Deployment of Future Mobility Technologies at the University of Warwick. Mehrdad Dianati holds part-time professorial posts at the School of Electronics, Electrical Engineering and Computer Science (EEECS), Queen’s University of Belfast and WMG at the University of Warwick. Other authors are with WMG, University of Warwick, e-mail: \{hakan.yatbaz, m.dianati, konstantinos.koufos, r.woodman\}@warwick.ac.uk,  m.dianati@qub.ac.uk}}



\maketitle

\begin{abstract}

Reliable detection of various objects and road users in the surrounding environment is crucial for the safe operation of automated driving systems (ADS). Despite recent progresses in developing highly accurate object detectors based on Deep Neural Networks (DNNs), they still remain prone to detection errors, which can lead to fatal consequences in safety-critical applications such as ADS. An effective remedy to this problem is to equip the system with run-time monitoring, named as \textit{introspection} in the context of autonomous systems. Motivated by this, we introduce a novel introspection solution, which operates at the frame level for DNN-based 2D object detection and leverages neural network activation patterns. The proposed approach pre-processes the neural activation patterns of the object detector's backbone using several different modes.  To provide extensive comparative analysis and fair comparison, we also adapt and implement several state-of-the-art (SOTA) introspection mechanisms for error detection in 2D object detection, using one-stage and two-stage object detectors evaluated on KITTI and BDD datasets. We compare the performance of the proposed solution in terms of error detection, adaptability to dataset shift, and, computational and memory resource requirements. Our performance evaluation shows that the proposed introspection solution outperforms SOTA methods, achieving an absolute reduction in the missed error ratio of 9\% to 17\% in the BDD dataset.

\end{abstract}

\section{Introduction}

\begin{figure}[t]
    \centering
    \includegraphics[width =0.9\linewidth]{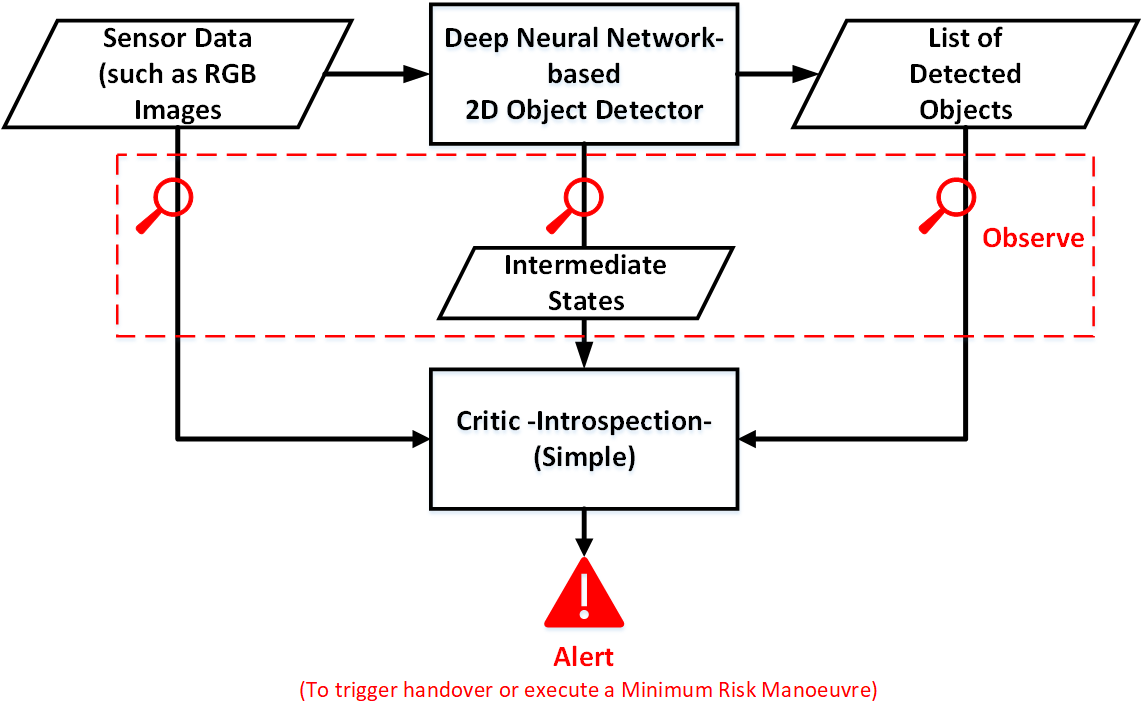}
    \caption{Actor-critic architecture for introspecting DNN-based ADS perception: Introspection can monitor perception input, intermediate model outputs, or the final output of the main system (or combinations of them). In case of an error, it should provide an alert to take further action such as handover or minimum risk manoeuvre \cite{koopman2016challenges}. Frame-level rather than object-level introspection is within the scope of this paper.}
    \label{fig:overview}
\end{figure}

\IEEEPARstart{T}{o} successfully operate in diverse driving scenarios, automated driving systems (ADS) must be able to perform three main functions, i.e., perception, planning, and control. Perception, which consists of sensing and perceiving the surrounding environment, is one of the most crucial operations, as the rest of the ADS architectural pipeline heavily depends on its ability to understand the environment. The core functionality of the perception system is to accurately recognise, i.e., localise and classify, other road users and surrounding objects in the driving scene. Failure to do so can lead to dangerous and costly road accidents. For example, in 2018, a Tesla vehicle collided with a roadside impact attenuator, leading to subsequent crashes and the death of its passenger~\cite{nhtsa:2018t}. The accident report, prepared by the National Traffic Safety Board, suggested that the lane detection algorithm momentarily failed due to excess sunlight affecting the vehicle's cameras. Such accidents highlight the critical need for perception systems to be resilient against errors that may arise, even when designed with the utmost care, in ADS. To ensure a fail-safe operation, ADS must be equipped with a tailored mechanism, conceptually depicted in Fig.~\ref{fig:overview}, to continuously monitor the perception system for potential errors \cite{hakansurvey}. Once such an error is detected, these mechanisms 
are expected to issue an alert which can be used to hand over the control to the driver in SAE Level 3 ADS or trigger a minimum risk manoeuvre in SAE Level 4 ADS \cite{sae}.

State-of-the-art (SOTA) object detection mechanisms often use deep neural networks (DNNs) that have recently shown remarkable performance on various benchmark datasets~\cite{detectionsurvey} and continue to be an active area of research for further improvement. Despite that, object detectors in ADS are not error-free for a number of reasons. First, the deployment conditions are so versatile that it is impossible to encompass all of them during training. Similarly, the training process is data-dependent and stochastic, e.g., the neural network weights may be randomly initialised. In other words, some parts of the DNNs are non-deterministic by design. Also, noise and input degradation can affect how well the DNN-based models work in practice~\cite{valentinanoise}. Hence, detecting errors in DNN-based systems fundamentally differs from traditional automation systems (e.g., anti-lock braking system) and is hereafter defined with a different term, \textit{``introspection"}.

An introspection mechanism may observe in run-time the input, intermediate states, and/or the output of the object detector as a critic system, see Fig.~\ref{fig:overview}. Most of the literature has so far focused on observing the model output for introspection, as it contains uncertainty/confidence indicators (the softmax values). For example, if the detected objects are associated with low confidence, this may suggest that their predictions are unreliable and an alert must be triggered.
However, the softmax values do not always represent realistic confidence in the model, and it has been shown that they can be misleading for introspection~\cite{dropoutod}. Furthermore, some recent studies have highlighted that it is possible to improve introspection using other methods and learning representations. For instance, in~\cite{rahmanper}, the intermediate states of object detection networks, also known as activation maps, are processed and used to detect erroneous predictions based on the model's performance on a selected metric, such as the mean average precision (mAP). Alternatively, in~\cite{fnyang}, the input image is directly utilised and fed into an object detector-like architecture to predict missed objects. Also, in~\cite{posthoc2022}, confidence values are combined with image features, such as the entropy of the colour histogram to discover erroneous cases. 

There has been recently an effort to develop introspection mechanisms by analysing different elements of the perception pipeline. A prominent study,~\cite{ash}, on out-of-distribution detection for image classification problems claims that over-parameterised activations of the main mechanism might be confusing to identify images that do not belong to one of the known classes used during training. Despite all recent efforts to expand the SOTA beyond confidence-based introspection using softmax probabilities, non-confidence-based introspection techniques are still under-explored and lack a unified framework for comparing the performances and computational requirements of introspection mechanisms. 

To that end, this paper expands upon our preliminary work in~\cite{hakaniccv}, which utilised pre-processed neural network activation maps for introspection, a method originally introduced in \cite{ash} for out-of-distribution detection. As compared to~\cite{hakaniccv}, the following contributions have been  incorporated into the present study: We present a detailed comparative analysis of the proposed introspection method against several other learning representation schemes adopted in the literature. Unlike~\cite{hakaniccv}, the new introspection framework is designed to accommodate not only the neural network activation patterns but also the detector's input and output. Its performance is extensively tested across several more object detectors and datasets, ensuring a robust and comprehensive evaluation of its applicability and effectiveness. Furthermore, we examine the computational and memory demands as well as the  performance under dataset-shift scenarios, aspects not covered  in~\cite{hakaniccv}.
To gain deeper understanding, we have adapted three SOTA mechanism where a new learning representation for introspection of 2D object detection is introduced. In addition to the comparison with SOTA, we have also utilise no processing as a separate learning representation for investigating the effect of performing the proposed mechanism for introspection. All of these learning representations, i.e., methods, are created by employing both one- and two-stage object detectors in our analysis, namely the fully-convolutional one-stage (FCOS)~\cite{fcos}, YOLOv8 \cite{yolo} and the Faster-RCNN~\cite{fasterrcnn} detectors. Similarly, two popular driving datasets are used for evaluation: KITTI~\cite{kitti} and Berkeley deep drive (BDD)~\cite{bdd100k}. In addition, the dataset-shift performance of each model is included to demonstrate its ability to generalise in unknown deployment scenarios.

In summary, the contributions of this study are:
\begin{itemize}
    \item A novel introspection method for 2D object detection in ADS is designed. It is based upon the pre-processing of raw activation maps from the last layer of its backbone neural network as the learning representation.
    \item A unified four-stage framework is introduced for introspection training and performance evaluation in terms of error detection capability.
        \item A comprehensive evaluation of the proposed mechanism is carried out encompassing several modes and learning representations for introspecting 2D object detection on two well-known driving datasets (KITTI and BDD) and three SOTA object detection models (FCOS, Faster-RCNN and YOLOv8).
    \item A comparative performance evaluation against several SOTA introspection methods is presented based on error detection accuracy, adaptability to dataset change (dataset shift, or operational domain shift), and computational efficiency in terms of memory and time consumption.
\end{itemize} 

The rest of this paper is organised as follows. Section~\ref{sec:relatedwork} provides a literature review on introspection methods for object detection in ADS. The proposed introspection method is presented in Section~\ref{sec:setup}. The selection of 2D object detectors, datasets, and key performance indicators is given in Section~\ref{sec:expres}. Over there, the SOTA introspection models and their performance comparison with the proposed scheme are presented. Section~\ref{sec:disc} discusses the results and highlights pros and cons of the models in terms of performance and complexity. Concluding remarks are provided in Section~\ref{sec:conclusion}.

\section{Related Work}\label{sec:relatedwork}
This section reviews the literature on introspection methods for object detection in ADSs, categorised into four primary areas.

\textbf{Confidence/Uncertainty-based Introspection:} This category includes methods that assess uncertainty in object detection to flag potential misdetections. Harakeh \textit{et al.} employ a Bayesian object detector with Monte Carlo Dropout (MCD) to represent location uncertainty and improve detection reliability by replacing traditional non-maximum suppression with Bayesian inference~\cite{bayesod, mcdropout}. Miller \textit{et al.} extend MCD to provide uncertainty values for label and bounding box predictions, enhancing single-shot detector performance in open-set conditions~\cite{dropoutod, ssd}. Other approaches like GMM-Det utilize Gaussian Mixture Models to measure uncertainty and establish a find-and-reject mechanism for open-set faults~\cite{gmmdet}. Further, methods such as PixelInv and STUD focus on identifying undetected objects due to environmental factors or out-of-distribution samples~\cite{pixelinv, du2022unknown}. For 3D LIDAR point clouds, some studies test various confidence/uncertainty mechanisms to enhance detection accuracy~\cite{9922026, 9665821}.

\textbf{Performance Metric-based Introspection:} These methods aim to detect performance drops by analyzing key metrics. Notably, the mean average precision (mAP) is often used to capture the model's object detection ability across different classes. Techniques like those in~\cite{rahmanper} employ neural networks to monitor variations in mAP, using the output of a convolutional neural network to extract features indicative of performance. Extensions of this approach involve monitoring sequences of frames or employing cascaded networks to enhance detection of performance drops~\cite{rahmancascade}. Additionally, Yang \textit{et al.} propose a method to predict object-level false negatives using an introspection model independent of the underlying object detector~\cite{fnyang}.

\textbf{Inconsistency-based Introspection:} Leveraging the multi-modality of ADS perception, these methods declare faults when outputs from different algorithms or sensors don't match. Techniques vary from using stereo and temporal cues to graph-based methods for identifying inconsistencies between processors or data inputs~\cite{failingtolearn, kdiagnose, antonante2022monitoring}.

\textbf{Past Experience-based Introspection:} Methods in this category utilize historical data to correlate detection performance with environmental characteristics. Systems like the one proposed by Hawke \textit{et al.} retrain networks with past false-negative samples to enhance performance. Other approaches employ location-specific methods, granting autonomy based on reliable past performance records or visual similarity-based experiences~\cite{Hawke2016, fitforpurpose, learnfromexp}.

\section{Proposed Introspection Model}\label{sec:setup}

\begin{figure*}[t]
    \centering
    \includegraphics[width = \linewidth]{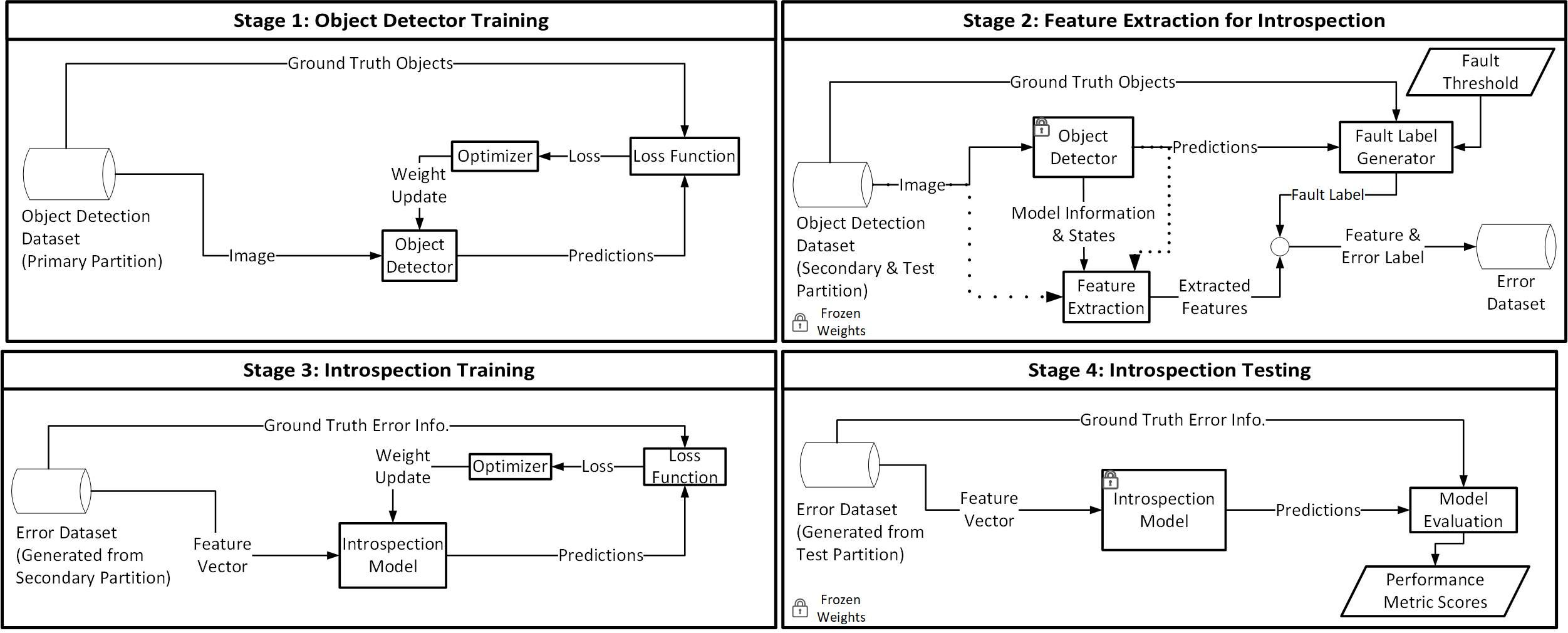}
    \caption{Four-stage framework for the comparative analysis of introspection models: (1) training an object detection model specifically for driving scenarios, diverging from generic pre-training datasets such as COCO and Pascal VOC. (2) Generation of an error dataset associating the features and labels for introspection. (3) Training the introspection system
using the error dataset from the validation set, and (4) evaluating the introspection system’s performance using the error dataset from the test set, with a corresponding feature, label pair. The dotted lines in the top right figure indicate that the input image and the output of the object detector might not be used for feature extraction by some of the introspection models (details will be provided in the description of each model).}
    \label{fig:framework}
\end{figure*}

This section introduces a new method for introspecting 2D object detection per-image using shaping of raw activation maps. Before that, we present a unified four-stage framework for the training and testing of performance metric-based introspection models, see Fig.~\ref{fig:framework}. That framework will also simplify the comparison of different introspection models in Section~IV.
\begin{figure}[t]
     \centering
     \includegraphics[width=\linewidth]{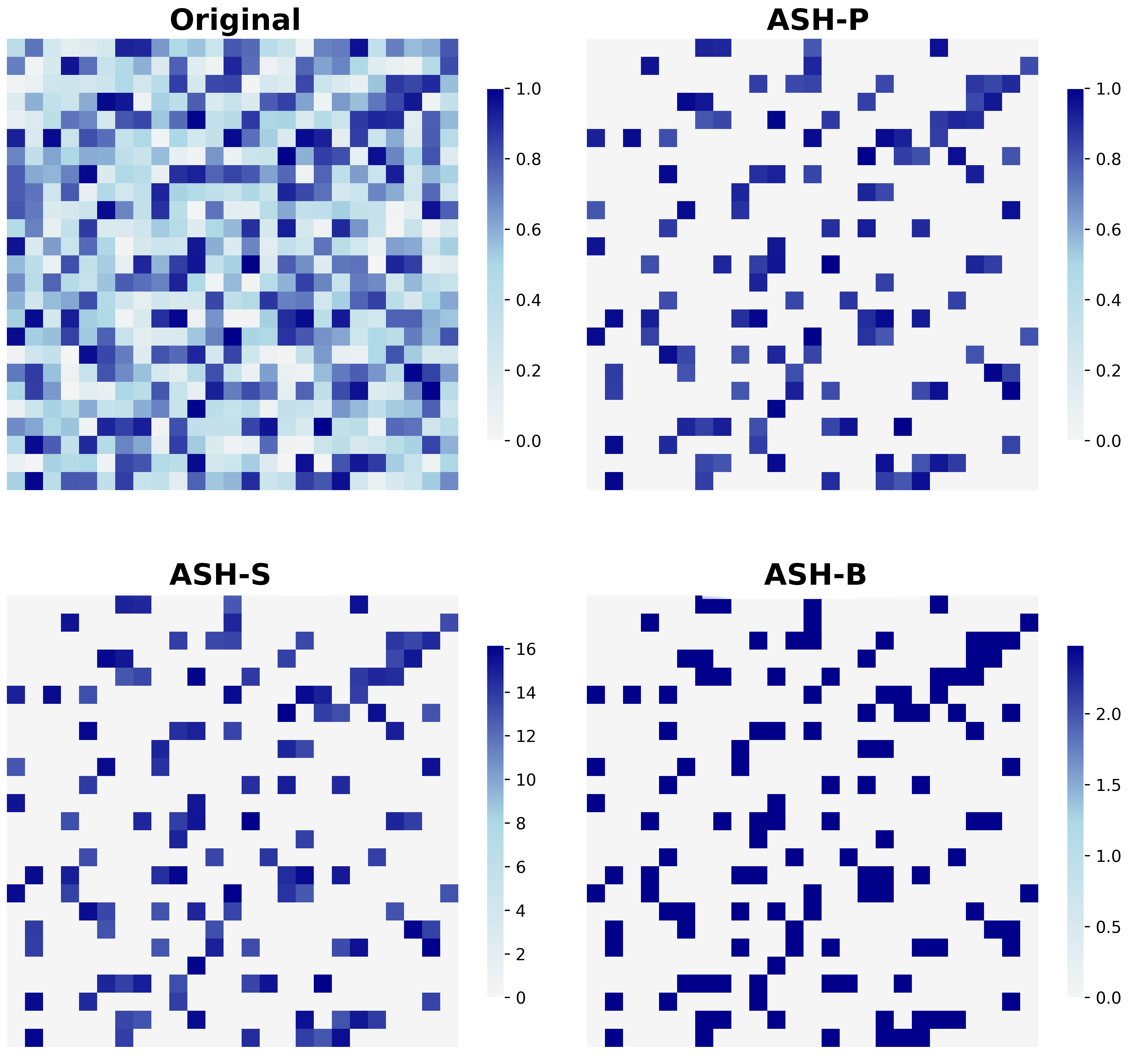}
     \caption{Example visualisations of activation shaping techniques with a selected scaling parameter $p=80$\%. `Original' depicts unaltered neural activation maps from the backbone network. 'ASH-P' retains the original activation scale without modification, simply pruning 80\% of the 'Original' map, ensuring a direct comparison, see Eq.~(1). 'ASH-S' shows scaled activations after pruning, with the color scale adjusted to reflect redistributed activation intensities. `ASH-B' represents binarised activations, with the scale indicating binary states of activation, diverging from the continuous values in 'Original', 'ASH-S' and 'ASH-P'. Each mode employs a different processing strategy, resulting in distinct scaling values, despite the similar visual patterns.}
     \label{fig:proposedmethod}
 \end{figure}
\begin{figure*}
    \centering
    \includegraphics[width=0.85\textwidth]{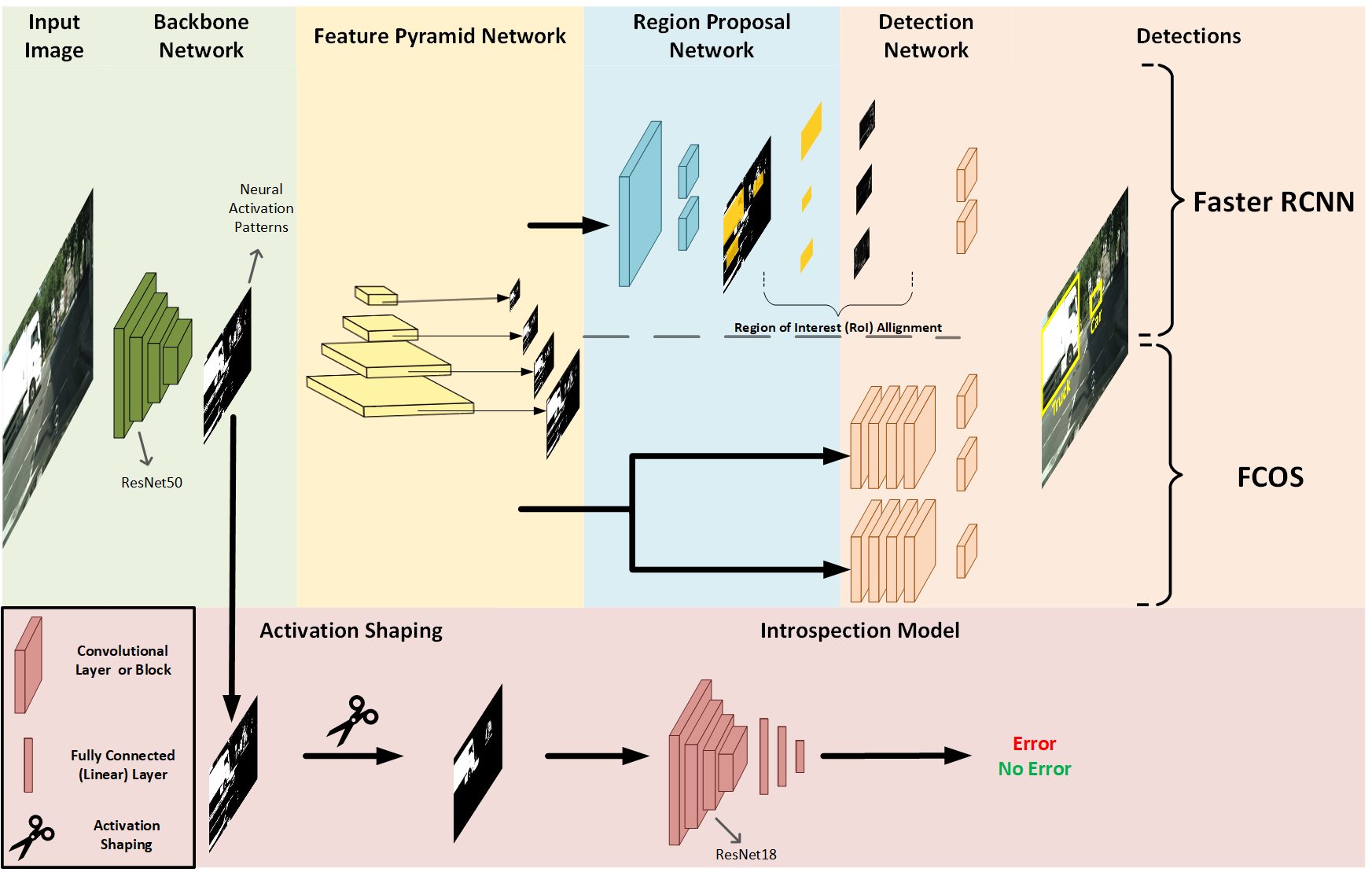}
    \caption{Architecture of the proposed mechanism in run-time: The top multi-coloured section illustrates the commonalities between the object detectors, including the backbone and Feature Pyramid Network (FPN) components. The distinction emerges post-FPN: The top line highlights the Faster-RCNN's utilisation of a Region Proposal Network (RPN) and Region of Interest (RoI) alignment for feature map standardisation before detection, while the bottom line emphasises FCOS's direct use of multi-scale feature maps to its detection block. Despite these differences, both methods employ a backbone structure from which our introspection mechanism requires access and extracts neural activation patterns. These patterns are then shaped before insertion into the introspection model for identifying errors. Finally, with reference to Fig.~\ref{fig:framework}, ResNet50 is fine-tuned during Stage~1, while  ResNet18 and the fully-connected network are trained during Stage~3 of the unified four-stage framework for the training and testing of metric-based introspection models. }
    \label{fig:systemdiagram}
\end{figure*}

With reference to Fig.~\ref{fig:framework}, we start by partitioning the driving dataset into training, test, and validation sets, with a ratio of 60-20-20\%, respectively. In the first stage, we train the object detection model as most detectors are pre-trained on generic datasets such as COCO~\cite{coco} or Pascal VOC~\cite{pascalvoc}. In the second stage, we use the validation and test sets to generate an error dataset for each introspection mechanism. In doing so, we extract and process information from different parts of the object detection pipeline and pair them with corresponding labels. To label the error dataset, we calculate the mAP and quantise it using predefined threshold value(s). Finally, in the third and fourth stages, we use the part of the error dataset generated by the validation/test set of the driving dataset to train/evaluate the introspection mechanism.

A recent study on OOD detection has highlighted that raw neural activation patterns can be confusing when the system needs to identify images containing objects that do not belong to one of the known classes used during training~\cite{ash}. This study also shows that simplification of activation patterns may help identify OOD samples without suffering from significant performance loss for the in-distribution samples. That observation motivates the investigation of the effect of pre-processing the neural activation patterns for in-distribution error detection.

To test our hypothesis, we extract the raw activation maps from the last layer of the backbone network of the object detector (pre-trained ResNet50) and associate them with the calculated mAP during the training phase (see Stage~2 in Fig.~\ref{fig:framework}). We have selected the mAP among other performance-based metrics, as this is the most commonly used metric for object detection. Similar to~\cite{rahmanper}, we use a threshold value of 0.5 for the mAP to label the (binary) error dataset, i.e., 'error' or 'no error'. Discussion on the effects of the value of the mAP threshold on the error detection performance will follow in Section~IV-G. The labelling of the error dataset is carried out on a per-frame basis. 
During the training and testing phases (see Stage~3 and Stage~4 in Fig.~\ref{fig:framework}), we pre-process the raw activation patterns before feeding them into the introspection model. Specifically, we follow the two-stage approach \cite{ash}, consisting of the following (see Fig.~\ref{fig:proposedmethod}):
\begin{enumerate} 
    \item Set equal to zero the activation elements whose values are less than the $p$-th percentile of the sample, i.e., 
    \begin{equation}\label{eq:prun}
   \mathbf{x}'= {\textnormal{Shape}}(\mathbf{x}) =
        \begin{cases}
        x_{i}, & \text{if } x_{i} \geq {F}^{-1}(p) \\
        0, & \text{otherwise},
        \end{cases}
    \end{equation}
    where $\mathbf{x}$ is the activation pattern, $x_{i}, i=1,\ldots, n$ is its $i$-th element, $F^{-1}$ is the inverse (empirical) cumulative distribution function of the activation pattern, and $\mathbf{x}'$ is the shaped activation pattern. 
    
    At the top row of Fig.~3, one may find an example illustration for the original activation map (left) and its shaped pattern (right) after eliminating, i.e., setting equal to zero, $80$\% of its elements (those associated with the lower values), while the remaining $20$\% of elements retain their original values.
    \item Process the remaining activations using one of the following rules:
    \begin{itemize}
        \item Keep the remaining activations as it is, called activation shaping with pruning \textbf{(ASH-P)}.
        \item Set all the values to a positive constant $\beta$ calculated using the sum of all activations divided by the number of unpruned activations called \textbf{ASH by binarisation (ASH-B)}.
        \begin{equation}\label{eq:binarize}
                \mathbf{x}' =
        \begin{cases}
        \beta, & \text{if } x_{i} \geq {F}^{-1}(p) \\
        0, & \text{otherwise}, 
        \end{cases}
        \end{equation}
        where $\beta=\frac{1}{|\{ i:x_i'\neq 0\}|}\sum_i x_i$, $x_i'$ is an element of the shaped activation pattern, and $\lvert \, \rvert$ stands for the cardinality, i.e., the number of elements in a set. Clearly, at the bottom-right corner in Fig.~3, an element is either zero or equals a constant $\beta$.
        \item Scale up all the activations by the ratio calculated with the sum of the activations before and after pruning, called \textbf{ASH with scaling (ASH-S)}.
        \begin{equation}\label{eq:scale}
                \mathbf{x}' =
        \begin{cases}
        \beta \, x_i, & \text{if } x_{i} \geq {F}^{-1}(p) \\
        0, & \text{otherwise},  
        \end{cases}
        \end{equation}
    \end{itemize}
\end{enumerate}
where $\beta = \exp\left( \frac{\sum_i x_i}{\sum_i x_i'}\right)$. It is emphasised that the value of the constant  $\beta$ for ASH-S is different than that calculated for ASH-B, see also the bottom-left shaped map in Fig.~3.


Note that for this method, the input to the feature extractor in Stage~2 is only the shaped activation patterns, i.e., the input image and the output of the object detector are not used for introspection. Also, the introspection model in Stage~3 and Stage~4  consists of a CNN (ResNet18) followed by a fully connected neural network (FCN). This indicates that the proposed mechanism assumes the ability to access the detector's backbone network for obtaining such activation maps. This is also evident from Fig.~\ref{fig:systemdiagram} where it becomes clear that during run-time the introspection mechanism is agnostic to the operation details of the object detector provided that it can access its activation maps.

Before continuing with the performance evaluation section, we note that the three schemes presented in this section are also referred to as learned features with activation shaping (LF-ASH). We will also consider the introspection framework without shaping the activation maps, which is hereafter referred to as learned features with raw activations (LFR).



\section{Performance Evaluations}\label{sec:expres}
In this section we first justify the selection of object detectors, driving datasets, selected SOTA introspection mechanisms and key performance indicators used in our analysis.
After that, we proceed with the experimental setup and the performance evaluation of the proposed introspection mechanism in comparison with SOTA models.

\subsection{Object Detectors}
We investigate the behaviour of introspection systems on 2D object detection using two popular object detection models. The former, Faster-RCNN~\cite{fasterrcnn}, is a widely used two-stage object detector in the computer vision domain, serving as a baseline for comparison. Faster-RCNN uses an additional neural network stage to propose regions with objects without any category information, which improves the performance but increases the complexity and required inference time. In recent years, there has been growing interest in one-stage object detectors, because of their ability to achieve fast inference times, despite typically having lower performance than two-stage detectors. To evaluate the performance of one-stage object detectors, we use the fully-convolutional one-stage (FCOS)~\cite{fcos} object detector that has shown promising performance in terms of speed and detection. In all experiments, we use the ResNet50 architecture as the backbone network for feature extractors for both object detectors.

In addition, to investigate the applicability of the proposed mechanism for real-life applications, we have utilised the YOLOv8 model~\cite{yolo}. YOLOv8 is the latest iteration in the YOLO series of real-time object detection systems. This model is known for its ability to perform object detection tasks rapidly and accurately by processing images in a single pass, hence the name "You Only Look Once. (YOLO)". It makes predictions with a single network evaluation. It utilises a different backbone network than FCOS and Faster-RCNN, which have ResNet backbones.  YOLOv8's architecture is designed to be both efficient and powerful, making it well-suited for scenarios where both speed and accuracy are critical such as ADS. Furthermore, a YOLO model has been utilised within the perception stack of the Autoware Foundation's open-source project for ADS \cite{autoware}.

\subsection{Driving Datasets}
To train and develop object detectors for ADS, there are several options available from companies and research institutes in the field. In this study, we select two popular datasets based on their use in both introspection and ADS domains: KITTI~\cite{kitti} and Berkeley Deep Drive (BDD)~\cite{bdd100k}. 

The KITTI dataset consists of over 14,000 annotated images captured by a camera and a Velodyne LiDAR mounted on a car driving through urban environments in Karlsruhe, Germany. The training set contains 7,481 annotated images with annotations for various object classes, and the test set includes 7,518 images. However, the benchmarking is typically done using only three classes: car, pedestrian, and cyclist. It's worth noting that the labels for the test set are not publicly available for fair benchmarking via the dataset's website.
The BDD dataset includes 100,000 annotated images taken from videos recorded in different parts of the USA. The dataset features ten classes, including car, pedestrian, bicycle, and motorcycle, but does not include the cyclist class. The 100,000 images are split into train, test, and validation sets with proportions of 70\%, 20\%, and 10\%, respectively. The test set does not have publicly available labels.

To ensure compatibility between the datasets, we merge the object classes into two main categories: vehicle and people. All vehicle types are re-labelled as "vehicle," and all classes of people that walk, stand, or sit down are merged into the "people" class. This simplification allows a more direct comparison between the two datasets in our experiments.
Since the two datasets differ in scale, they result in imbalanced error datasets due to varying numbers of errors and diverse road traffic conditions. To mitigate this issue, we incorporate a class-weighted loss function during training. This adjusts the imbalance in the dataset by scaling up the loss for the minority class and scaling down the loss for the majority. The calculation of the weights for the $c$-th class can be read as 
\begin{equation}
        W(c) =  \frac{n_S}{n_{C}\cdot n_{c,s}},
        \label{eq:cweight}
\end{equation} 
where, $n_S$ is the total number of samples in the training set, $n_C$ is the total number of classes, and $n_{c,s}$ is the total number of samples in the selected class~\cite{classweight}.

\subsection{Adapted SOTA Introspection Mechanisms}
For comparison purposes, we implement three SOTA introspection mechanisms for object detection proposed in previous studies \cite{rahmanper, rahmancascade, posthoc2022}, which demonstrate strong performance in the field. Each of these mechanisms, named after their respective learning representations, has a distinct approach to error detection and involves different neural network architectures. For example, while the proposed scheme in Section~III requires a CNN for feature extraction prior to the classification mechanism, the studies in this section can directly feed the learning representation to an FCN. Therefore, we hereafter focus on the learning representations proposed by each study. 

\subsubsection{Statistical Features (SF)}
\begin{figure}[t]
    \centering
    \includegraphics[width=\linewidth]{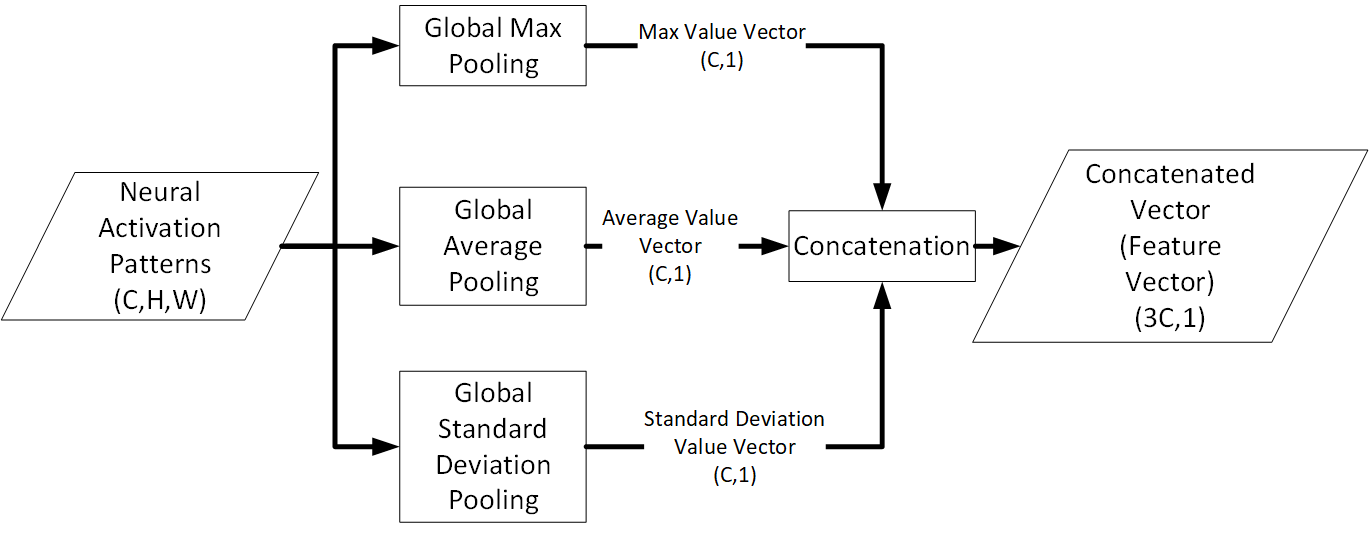}
    \caption{Illustration of the feature extraction process from ~\cite{rahmanper}. The 3D activation maps undergo global pooling operations (mean, max, and standard deviation) across their height and width. The processed 1D vectors are concatenated to form a unified column vector, serving as the learning representation titled as Statistical Features (SF).}
    \label{fig:rahmanper}
\end{figure}

The authors in~\cite{rahmanper} propose a mechanism that utilises the activation maps of the object detector's backbone CNN to extract features for error detection. Similar to LF-ASH, the activations are extracted from the last layer of pre-trained ResNet50. Specifically, the activation maps are generated by convolving the input image with multiple kernels, and the resulting responses of each neuron are captured in the feature maps. These 3D maps provide a comprehensive view of the learned features and can assist in interpreting the neural network's response to the input image. The authors apply mean, max, and standard deviation (std) functions on the activation maps, which are originally three-dimensional (3D), $\text{height}\times \text{width} \times \text{channels}$ or $H\times W\times C$. To generate and convert the learning representation into a 1D vector, global pooling is applied across height and width. The resulting vectors are concatenated into a column vector, which is used to train a multi-layer perceptron for binary classification, i.e., 'error' or 'no-error'. Fig.~\ref{fig:rahmanper} provides an illustration of the statistical feature extraction,   

\subsubsection{Cascaded Learned Features (CLF)}
\begin{figure}[t]
    \centering
    \includegraphics[width=\linewidth]{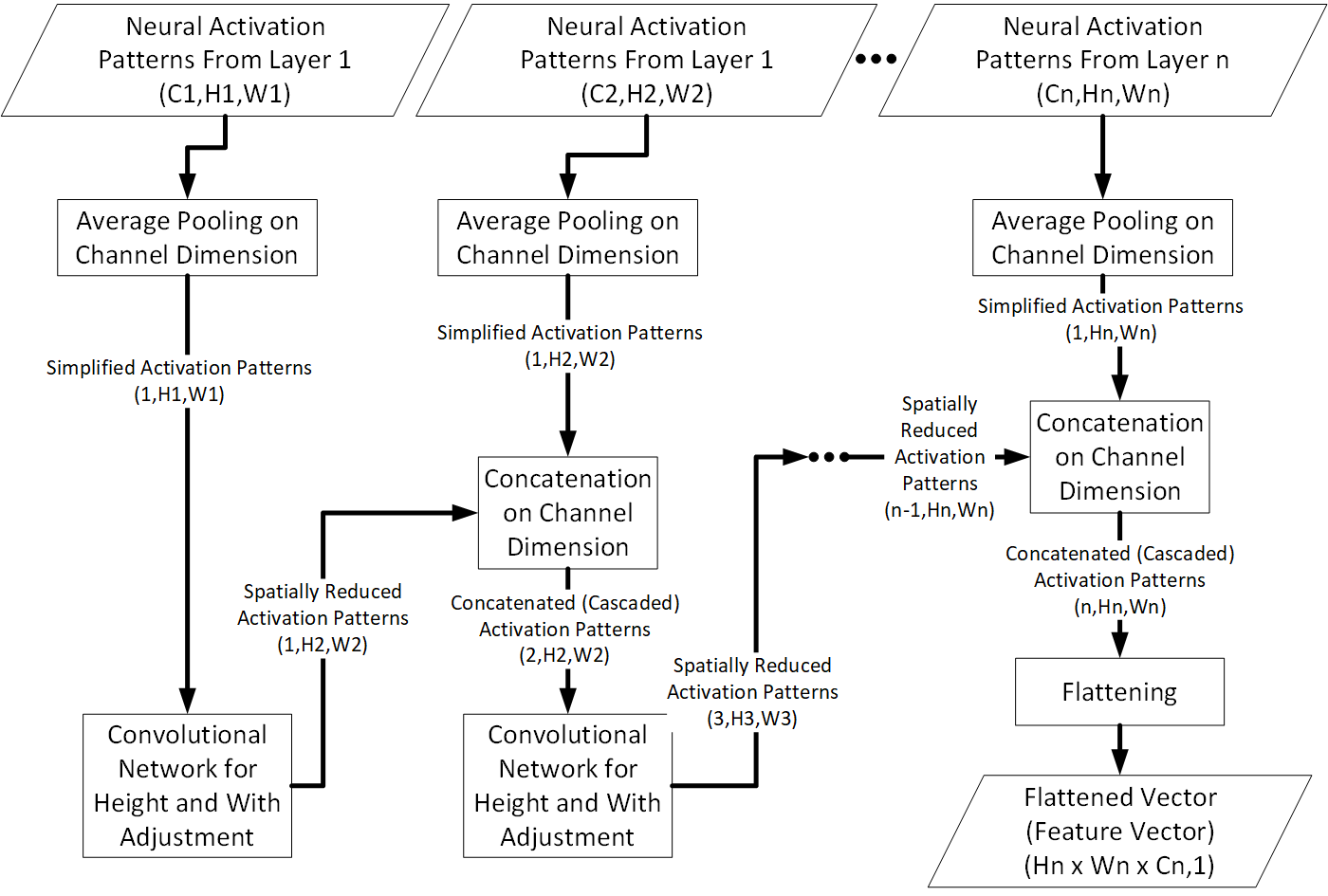}
    \caption{This figure illustrates an adapted cascade-based architecture, as referenced from ~\cite{rahmancascade}, designed to generate a composite learning representation by leveraging multiple layers of 2D object detector's backbone. Initially, 3D activation maps are extracted from these layers, which are then condensed into 2D activation maps via channel-wise average pooling. Subsequent processing through a convolutional network adjusts the dimensions of these maps before they are concatenated with similarly processed activations from successive layers. The resulting flattened vector termed the Cascaded Learned Features (CLF) represents the final distilled learning representation.}
    \label{fig:cascade}
\end{figure}

Rahman \textit{et al.} in~\cite{rahmancascade} extend their previous study in~\cite{rahmanper} to utilise activation maps from multiple layers of the detector's backbone to generate the proposed learning representation. The authors achieve this by iteratively concatenating multiple layers and applying pooling and convolutions in between to extract better features. However, it is worth noting that their system incorporates temporal data, which is not within the scope of our study. Therefore, we need to adapt the architecture of their cascaded network to remove the temporal concatenation component. The resulting cascade-based architecture is illustrated in Fig.~\ref{fig:cascade} for clarity. This mechanism which utilises convolutions on different activation maps and a cascade operation, is hereafter referred to as cascaded learned features (CLF).

Besides the temporal aspect, the CLF model is structured around ordinal classification, i.e., multi-class classification with the sense of ordering, but not numerical relation. While the majority of the studies perform binary classification indicating either 'error' or 'no-error', the reason for using this study as one of the baselines is to illustrate the effect of extracting and cascading features from the backbone on introspection. Similar to LFR and SF models, a mAP score equal to 0.5 is used to generate the (binary) error dataset. Also, note that neither SF nor CLF utilises the input image and the output of the object detector for feature extraction.
\subsubsection{Handcrafted Image \& Model Features (HIMF)}
In contrast to the previous methods, Zhang \textit{et al.} in~\cite{posthoc2022} propose a direct regression model to predict errors using both raw input features from the image and features extracted from the predictions. The authors define two sets of features: (i) image features, and (ii) model features. Image features consist of the image's colour histogram entropy, image size, and the number of corners in the image. Model features consist of three sets of information. First, class and location scores of the predictions are calculated, indicating the importance of a class or location. Second, minimum, maximum, and mean confidence values across all predictions are calculated. Third, the number of bounding boxes, minimum and mean bounding box sizes across subjects are found and used as handcrafted model features. Similar to the CLF method, the proposed feature extraction method is coupled with a binary classification model following the neural network architecture given in~\cite{posthoc2022}. Note that the original study focuses on regressing the selected metric rather than classifying, hence their mechanism is adapted for classification.

\subsection{Performance Metrics}\label{sec:metrics}
Since the decision of the introspection method is based on binary classification for all models considered in this paper, the following metrics are selected to evaluate the performance. 
\begin{itemize}
    \item \textbf{F1 Score:} Harmonic mean of precision and recall metrics. This metric is selected due to the imbalanced nature of the KITTI dataset.
    \begin{equation}
            \text{F1 Score}=\frac{2\cdot \text{Recall} \times \text{Precision}}{\text{Recall}+\text{Precision}}.
    \end{equation}
    \item \textbf{Area Under Receiver Operating Characteristic Curve (AUROC): } Provides an indicator of how well a classifier distinguishes between the positive ('error') and negative ('no-error') class.
    \item \textbf{False Negative Rate (FNR):} Indicates the ratio of cases where the introspection system could not detect the fault example, divided by the total number of fault cases in the test part of the error dataset. This metric quantifies the ratio of missed faults in the system, which is paramount as the consequences of false negatives can be severe.
    \begin{equation}
        \text{FNR} = \frac{\text{False Negatives}}{\text{False Negatives} + \text{True Positives}}.
    \end{equation}
\end{itemize}

Furthermore, it is essential to highlight that there is a class imbalance and a low number of error samples for experimentation on the KITTI dataset. This is due to data partitioning prior to the four-stage framework, and the overall number of samples available in the KITTI dataset, which is low. Hence, the AUROC metric can provide high values if the model is able to correctly identify non-erroneous samples that constitute the majority class. To better understand the model performance in such cases, we have also examined the FNR value to ensure that the model can sufficiently identify both erroneous and safe cases. On the contrary, due to the diversity and higher number of samples in the BDD dataset, there is no significant imbalance in either the training or testing set for introspection.

\subsection{Comparative Performance Evaluation between Different Pre-processing Modes}

To find the best-performing pre-processing mode, we first evaluated each mode presented in Section~III on both datasets with FCOS object detector. Since common ADSs utilise one-stage detectors, we opt to evaluate the presented modes only on the FCOS model. In addition, to optimise the introspection performance, we have extensively evaluated the performance for various combinations of hyperparameters, i.e., batch size, learning rate and focusing parameter $\gamma$. Their best values for each dataset, pre-processing mode and removal percentiles ranging from 70\% to 90\% can be retrieved from Table~\ref{tab:modeparams}.  The above-mentioned imbalance problem can also be seen in the parameter setups for best-performing models, especially considering the $\gamma$ value. Specifically, for the KITTI dataset, better results are obtained with higher $\gamma$ values in which the dominance of the majority samples is reduced.
\begin{table}[t]
\begin{center}
\caption{Best performing hyperparameters for each mode (S and P).}
\label{tab:modeparams}
\begin{tabular}{cccccc}
\textbf{Dataset} & \textbf{Mode} & \textbf{Percentile} & \makecell{\textbf{Batch}\\\textbf{Size}} &  $\gamma$ &\makecell{\textbf{Learning}\\\textbf{Rate}} \\ \toprule
\multirow{10}{*}{BDD} & \multirow{4}{*}{S} & 90   & 16  & 0 & 0.005 \\
      &             & 85  & 16  & 5 & 0.001 \\
      &             & 80  &  16   & 5 & 0.001 \\
      &             & 75  & 16   & 0 & 0.005  \\
      &             & 70    &  16    &  0  &  0.005    \\\cmidrule{2-6}
      &\multirow{4}{*}{P} & 90   & 16   & 5 & 0.010  \\
      &             & 85  & 16   & 0 & 0.005  \\
      &             & 80 & 16   & 0 & 0.001  \\
      &             & 75  & 16   & 0 & 0.005 \\
      &             & 70    & 16     & 0  & 0.001     \\\midrule
\multirow{10}{*}{KITTI} & \multirow{4}{*}{S}        & 90   & 16   & 0 & 0.005 \\
      &             & 85  & 32   & 0 & 0.001  \\
      &             & 80  & 64  & 5 & 0.010 \\
      &             & 75  & 64   & 5 & 0.005 \\
      &             & 70  & 64    & 5  & 0.001     \\\cmidrule{2-6}
      & \multirow{4}{*}{P} & 90   & 64   & 5 & 0.010 \\
      &             & 85  & 128   & 5 & 0.010  \\
      &             & 80 & 128   & 5 & 0.005 \\
      &             & 75  & 128   & 5 & 0.010\\ 
      &             & 70    &   128   & 5  &  0.005    \\\bottomrule
       
\end{tabular}
\end{center}
\end{table}

 The performance evaluation for both datasets is presented in Table~\ref{tab:moderes} for the pre-processing modes: only pruning (P), and pruning with scaling (S), where it is demonstrated that only pruning yields the best overall performance. This is in contrast with the outcome of the numerical results obtained in~\cite{ash} where it is shown that mode S outperforms mode P. This difference may be attributed to the varying use of pre-preprocessed activation patterns in our study and \cite{ash}. In~\cite{ash}, these patterns are used to calculate a different value, called energy score~\cite{energy}, while our introspection method utilises them for feature extraction and learning patterns for error detection. 
 
 For the BDD dataset, pruning alone (P) produces consistent results, although the FNR still varies between 0.11 and 0.35. On the contrary, it is apparent that the  FNR significantly fluctuates when scaling is applied after pruning (S). Overall, S achieves an AUROC of 0.76-0.80, indicating good performance. Nonetheless, considering also the FNR metric, we observe that high AUROC values are accompanied by higher FNR values such as 33\%. This behaviour indicates that the model tends to provide better performance for the not error cases as compared to the erroneous cases.

In the KITTI dataset, we observe a similar performance pattern as in the BDD dataset for the S mode, where results tend to lean towards one of the classes, e.g., on the one hand, a model with AUROC/FNR equal to 0.8088/0.7144 correctly identifies mostly non-erroneous cases, while on the other hand, a model with AUROC/FNR equal to 0.4584/0.0102 correctly detects mainly the error cases. Additionally, due to the limited number of samples in comparison with the BDD dataset, the performance inconsistencies between different percentiles are more pronounced in the KITTI dataset. 
\begin{table}[t]
\begin{center}
\caption{Comparison of the different pre-processing modes (S and P).}
\label{tab:moderes}
\begin{tabular}{ccccc}
\textbf{Dataset}&\textbf{Mode}                          & \makecell{\textbf{Percentile}} & \textbf{AUROC} & \textbf{FNR}  \\ \toprule 

\multirow{10}{*}{BDD}&\multicolumn{1}{c}{\multirow{5}{*}{S}} & 90                  & 0.7994         & 0.3302       \\
&                   & 85                  & 0.8057         & 0.2996       \\
&                   & 80                  & 0.7612         & 0.0180        \\
&                   & 75                  & 0.8021         & 0.0952       \\
&                   & 70                  & 0.7971         & 0.1114       \\ \cmidrule{2-5}
&\multirow{5}{*}{P}                     & 90                  & 0.8009         & 0.2611       \\
&                                       & 85                  & 0.8068         & 0.3521       \\
&                                       & 80                  & 0.7972         & 0.2374       \\
&                                       & \textbf{75}         & \textbf{0.8103} & \textbf{0.1069}\\
 &                                      & 70                  & 0.7999         & 0.2306       \\\midrule
\multirow{10}{*}{KITTI}&\multirow{5}{*}{S} & 90                  & 0.8088         & 0.7143       \\
&                   & 85                  & 0.6759         & 0.2143       \\
&                   & 80                  & 0.4584         & 0.0102       \\
&                   & 75                  & 0.5362         & 0.2449       \\
&                   & 70                  & 0.6102         & 0.1327       \\ \cmidrule{2-5}
&\multirow{5}{*}{P} & \textbf{90}                  & \textbf{0.8409}          & \textbf{0.4898}       \\
&                   & 85                  & 0.8346       & 0.4082       \\
&                   & 80                  & 0.8238         & 0.4796       \\
&                   & 75                  & 0.8235         & 0.4592       \\
&                   & 70                  & 0.8330          & 0.4388       \\\bottomrule
\end{tabular}
\end{center}
\end{table}

\subsection{Comparative Performance Evaluation between Adapted Mechanisms}

In order to conduct an extensive comparison of the four introspection methods, we evaluate their performances on all possible combinations of driving datasets and object detectors presented earlier in this section. For completeness, we also include the performance evaluation results of the method proposed in Section~III but without pre-processing of the extracted activation patterns, which is referred to as Learned Features with Raw activations (LFR). Additionally, since the best-performing mode of LF-ASH is only pruning mode (ASH-P), the results presented for LF-ASH in the following sections are obtained using ASH-P mode. To ensure a fair comparison, we use fixed seeds to split the data for the training, validation and test sets, as presented in Fig.~\ref{fig:framework}. Additionally, we apply hyper-parameter tuning on various parameters, such as the gamma parameter in the focal loss function ($L(\mathbf{q})$ in Eq.~\ref{eq:focal})~\cite{focal}, the learning rate scheduler, and batch size, see Table~\ref{tab:params}. We omit the number of epochs, optimiser type, and patience values from the table since the best models are found to use the same values for these parameters. Specifically, we use 600 epochs, stochastic gradient descent (SGD) for the optimiser, and 25 for the early stop patience.

For completeness, the focal loss function used to train the object detectors (FCOS, Faster-RCNN and YOLOv8) can be read as~\cite{focal}

\begin{equation}\label{eq:focal}
     L(\mathbf{q}) = -\sum\nolimits_i \alpha_{i} (1 - q_{i})^\gamma \log(q_{i}) \quad i=0, 1,
\end{equation}
where $\log$ is the natural logarithm, $\mathbf{q}$ is the predicted probability vector for the classes no error (0) and error (1) with elements ($q_0$ and $q_1$ respectively), $\alpha_i$ is a scaling factor (class weights) that balances the contribution of the positive and negative examples for each class, and $\gamma \in \{0, 2, 4, 5\}$ is a focusing parameter that down-weights easy examples and emphasises hard examples.
\begin{table}[t]
\begin{center}
\caption{Best parameters obtained from hyperparameter tuning for each configuration of dataset, detector and introspection model.}
\label{tab:params}
\begin{tabular}{cccccc}
\textbf{Dataset} & \textbf{Detector} & \textbf{Method} & \makecell{\textbf{Batch}\\\textbf{Size}} & \makecell{\textbf{Focal}\\\textbf{Loss}\\\textbf{Gamma}} & \textbf{LR} \\ \toprule
\multirow{10}{*}{BDD} & \multirow{5}{*}{FCOS} & SF   & 32  & 0 & 0.001 \\
      &             & CLF  & 128  & 4 & 0.005 \\
      &             & HIMF & 64   & 4 & 0.005 \\
      &             & LFR  & 32   & 0 & 0.010 \\
      &             & LF-ASH& 16 &  0 & 0.005\\\cmidrule{2-6}
      &\multirow{5}{*}{F-RCNN} & SF   & 64   & 0 & 0.010 \\
      &             & CLF  & 64   & 4 & 0.010 \\
      &             & HIMF & 32   & 5 & 0.010 \\
      &             & LFR  & 32   & 2 & 0.005 \\
      &             & LF-ASH& 32 &  0 & 0.01 \\\midrule
\multirow{10}{*}{KITTI} & \multirow{5}{*}{FCOS}        & SF   & 64   & 5 & 0.001 \\
      &             & CLF  & 64   & 0 & 0.010 \\
      &             & HIMF & 32  & 5 & 0.001 \\
      &             & LFR  & 64   & 0 & 0.001 \\
      &             & LF-ASH& 64   & 5 & 0.010 \\\cmidrule{2-6}
      & \multirow{5}{*}{F-RCNN} & SF   & 64   & 0 & 0.005 \\
      &             & CLF  & 32   & 5 & 0.010 \\
      &             & HIMF & 32   & 5 & 0.001 \\
      &             & LFR  & 32   & 5 & 0.001 \\
      &             & LF-ASH &32 &  5 & 0.001 \\ \bottomrule
       
\end{tabular}\\
\vspace{0.2cm}
 \footnotesize{\textbf{LR:}} \footnotesize{Learning Rate}   \tab[0.3cm]\footnotesize{\textbf{F-RCNN: }}\footnotesize{Faster R-CNN} \\
\end{center}
\end{table}


\begin{table}[t]
\begin{center}
\caption{Performance evaluation of each introspection method on selected datasets and object detectors.}
\label{tab:performance}
\begin{tabular}{cccccc}
\textbf{Dataset} & \textbf{Detector} & \textbf{Method} & \textbf{AUROC} & \textbf{F1} & \textbf{FNR} \\ \toprule
\multirow{10}{*}{BDD}   & \multirow{5}{*}{FCOS}   & CLF    & 0.6921 & 0.6409 & 0.3335 \\
                        &                         & HIMF   & 0.6631 & 0.5168 & 0.9686 \\
                        &                         & LF-ASH & 0.8103 & 0.7394 & 0.1069 \\
                        &                         & LFR    & 0.7793 & 0.7423 & 0.2439 \\
                        &                         & SF     & 0.8365 & 0.7526 & 0.1983 \\ \cmidrule{2-6}
                        & \multirow{5}{*}{F-RCNN} & CLF    & 0.6966 & 0.6372 & 0.2724 \\
                        &                         & HIMF   & 0.6655 & 0.6158 & 0.4637 \\
                        &                         & LF-ASH & 0.7683 & 0.7239 & 0.1679 \\
                        &                         & LFR    & 0.7615 & 0.7155 & 0.2405 \\
                        &                         & SF     & 0.8250 & 0.7274 & 0.3496 \\  \midrule
\multirow{10}{*}{KITTI} & \multirow{5}{*}{FCOS}   & CLF    & 0.7615 & 0.8128 & 0.4490 \\
                        &                         & HIMF   & 0.6049 & 0.4017 & 0.2347 \\
                        &                         & LF-ASH & 0.8409 & 0.7098 & 0.4898 \\
                        &                         & LFR    & 0.8065 & 0.8596 & 0.3673 \\
                        &                         & SF     & 0.8315 & 0.8328 & 0.3163 \\ \cmidrule{2-6}
                        & \multirow{5}{*}{F-RCNN} & CLF    & 0.7406 & 0.7112 & 0.3458 \\
                        &                         & HIMF   & 0.5715 & 0.6471 & 0.5421 \\
                        &                         & LF-ASH & 0.8298 & 0.6252 & 0.3551 \\
                        &                         & LFR    & 0.8150 & 0.8048 & 0.3738 \\
                        &                         & SF     & 0.8401 & 0.8951 & 0.4299 \\ \bottomrule
\end{tabular}
\end{center}
\end{table}

\subsubsection{Detection performance}\label{sec:detperf} Table~\ref{tab:performance} displays the evaluation results of the considered introspection methods. Before delving into their performance comparison, we would like to clarify that the F1 score presented in this table is the average across the different classes. According to Table~\ref{tab:performance}, the SF model attains the best discriminatory performance with AUROC values varying from 83-84\%~\footnote{Rounding of the minimum and the maximum value to the nearest integer is used in the text.}, and F1 Scores ranging from 72-89\%. The LFR method provides competitive results with a range of 76-82\% for the AUROC and 72-86\% for the F1 Score, while activation shaping (LF-ASH) boosts up the performance to 76-84\% for AUROC but degrades the F1 Score to 62-73\%. The CLF method also achieves competitive performance yielding an AUROC score between 69-76\% and an F1 Score between 63-81\%. Lastly, we see that the HIMF method is not as capable as the other three methods attaining a maximum of 66\% and 64\% for the AUROC and F1 Score, respectively. Overall, the results show that SF representation outperforms the other methods in these two metrics and provides better error detection performance. 

Due to the repercussions of false negatives in the operation of safe-critical systems, it is essential that the missed detection errors are kept at a minimum level, even if this means that some degree of discomfort must be tolerated by the ADS, i.e., unwanted interruptions to the system operation. For that reason, the FNR metric (the lower, the better) is also reported in Table~\ref{tab:performance}. One can see that the CLF, LFR and LF-ASH are more reliable introspection methods for detecting errors than the SF mechanism under the Faster-RCNN detector. The opposite is true when the FCOS detector is employed except the proposed LF-ASH method. Moreover, the HIMF model has high/low FNR in the BDD/KITTI dataset. Considering the overall detection capability of HIMF, these results indicate that HIMF classifies most samples as one class, and it does not produce a sufficient learning representation for the subject application.


\subsubsection{Cross-dataset performance} The evaluation of object detection models on specific datasets is crucial for understanding their performance in real-world scenarios. However, the performance of these models can be significantly affected by changes in the distribution of the data between the training and testing stages, which is known as dataset shift. This is particularly relevant in the context of ADS, where the environment can change rapidly and unpredictably. In order to ensure the robustness and reliability of introspection systems, it is essential to evaluate their performances under various sources of dataset shift, such as changes in weather, or lighting conditions. Hence, we investigate the effect of dataset shift on the introspection models by evaluating the models trained on KITTI dataset with BDD dataset and vice versa. It should be noted that due to the low performance of the HIMF method, we exclude it from this experiment. 
\begin{table}[t]
\begin{center}
\caption{Performance evaluation of each introspection method on selected datasets and object detectors.}
\label{tab:shift}
\begin{tabular}{ccccccc}
\makecell{\textbf{Train}\\\textbf{Set}}& \makecell{\textbf{Test}\\\textbf{Set}} & \textbf{Detector} & \textbf{Method} & \textbf{AUROC} & \textbf{F1} & \textbf{FNR} \\ \toprule
\multirow{8}{*}{BDD}   & \multirow{8}{*}{KITTI} & \multirow{4}{*}{FCOS} & CLF & 0.5000 & 0.4944 & 0.0000 \\
   &  &  & LFR & 0.3612 & 0.5051 & 0.9991 \\ 
   &  &  & SF  & 0.6142 & 0.5727 & 0.2386 \\
   &  &  & LF-ASH & 0.4744 & 0.0600 & 0.0000 \\ \cmidrule{3-7}
   &  & \multirow{3}{*}{F-RCNN} & CLF & 0.3320 & 0.4334 & 0.8524 \\
   &  &  & LFR & 0.4828 & 0.5245 & 0.9985 \\
   &  &  & SF  & 0.5854 & 0.5510 & 0.8628 \\ 
   &  &  & LF-ASH & 0.6401 & 0.1271 & 0.0094 \\ \midrule
\multirow{8}{*}{KITTI} & \multirow{8}{*}{BDD}   & \multirow{4}{*}{FCOS}  & CLF & 0.4131 & 0.0655 & 0.0000 \\
 &  &  & LFR & 0.3775 & 0.0655 & 0.0000 \\
 &  &  & SF  & 0.5754 & 0.1197 & 0.0204 \\
 &  &  & LF-ASH & 0.5000 & 0.3200 & 0.0000 \\\cmidrule{3-7}
 &  & \multirow{4}{*}{F-RCNN} & CLF & 0.4333 & 0.8543 & 0.9439 \\
 &  &  & LFR & 0.5160 & 0.8630 & 0.9065 \\
 &  &  & SF  & 0.6571 & 0.3189 & 0.1308 \\ 
 &  &  & LF-ASH &0.4334 & 0.3115& 0.0104  \\ \bottomrule

\end{tabular}
\end{center}
\end{table}

The cross-dataset performance of four introspection methods is presented in Table~\ref{tab:shift}, where one can observe that the CLF, LFR and LF-ASH models have very limited ability to cope with dataset shift as compared to the SF model. The SF representation achieves 58-66\% AUROC, 12-58\% F1 Score and 2-86\% FNR values. Although the performance of the SF mechanism is inconsistent among datasets and detectors, the results in Table~\ref{tab:shift} highlight that some of its error detection ability is maintained across different environments/datasets. On the contrary, the CLF, LFR and LF-ASH methods provide low FNR associated with low AUROC and/or F1 Score values indicating that they classify the majority of samples to the error class. For example, the four introspection models using the FCOS detector and tested in the BDD dataset do not miss any errors but they also have very low AUROC and F1 Score. 

\subsubsection{Computational complexity} Besides the error detection performance, it is also important to consider the computational and memory requirements of the introspection models, particularly in the context of ADS, where the response time is crucial for fail-safe control. ADS have restricted resources available for computing and storage, which can significantly impact the feasibility of implementing sophisticated deep learning models for perception and introspection. Therefore, while the performance of introspection models is a crucial factor in determining their usefulness in ADS, it is equally important to ensure that these models can operate within limited available resources. To this end, we present the next quantitative results on the size of the generated input, memory usage on the GPU, and the time taken for each input representation on both the CPU and GPU during inference of one sample. The information is collected using PyTorch profiler and Pytorch cuda module on a device equipped with an Intel(R) Xeon(R) Silver 4216 CPU @ 2.10GHz and NVIDIA RTX 3090 24GB GPU.

Table~\ref{tab:comp} presents a comparison of the computational and memory requirements of the different learning representations. It is shown that the HIMF is the fastest and least memory-intensive representation, followed by the SF, which simplifies the neural network activation using statistical functions. The CLF and LF-ASH~\footnote{Note that in terms of storage requirements and inference time, there is no difference between raw and pre-processed activation patterns.} mechanisms contain a larger volume of information for error detection resulting in longer inference times and higher memory requirements. 
It is worth noting that the input sizes presented in Table~IV are not the outcomes of the most optimal saving mechanisms, but they can still illustrate the distinctions between different learning representations. Finally, the results in Table~V can be used to infer that the proposed method (LF-ASH) requires less combined time for pre-processing and inference than the HIMF method. 
\begin{table}[t]
\begin{center}
\caption{Inference time and memory requirements of each introspection method.}
\label{tab:comp}
\begin{tabular}{ccccc}
\textbf{Method} & \textbf{CPU Time}& \textbf{GPU Time} & \makecell{\textbf{Peak}\\\textbf{CUDA}\\\textbf{Memory}}&\textbf{Input Size} \\ 
& (s) & (ms) & (MB) & (MB) \\
\toprule
\textbf{CLF} & 3.162& 0.200 & 8.723 & 0.336 \\
\textbf{LF-ASF} & 3.185& 2.981& 77.92 & 7.9 \\
\textbf{HIMF} & 1.096& 0.018 & 0.003 & 0.004 \\
\textbf{SF} & 1.088& 0.036& 12.529 & 0.029  \\ \bottomrule

\end{tabular}
\end{center}
\end{table}
\begin{table}[t]
\begin{center}
\caption{Pre-processing time of each introspection method.}
\label{tab:preprocesscomp}
\begin{tabular}{ccccc}
\textbf{CLF} & \textbf{LFR} & \textbf{HIMF} & \textbf{SF} & \textbf{LF-ASH} \\ \toprule
N/A & N/A &203.51~ms & 8.33~ms & 74.66~ms \\ \bottomrule
\end{tabular}
\end{center}
\end{table}
\subsubsection{Summary} In a nutshell, the evaluation results of several introspection methods presented in this section indicate that the SF and the (proposed) LF-ASH methods provide the best error detection performances in terms of  AUROC and F1 Score values. The performance of the CLF method is lower but still competitive, while the HIMF method is not as capable as the other presented methods. Furthermore, the LF-ASH method outperforms the SF method regarding missed error detection ratios, which is paramount for safe-critical applications such as ADS. Therefore, we claim that the LF-ASH scheme delivers the best error detection performance when there is a larger set of training samples, such as in the case of the BDD dataset.Nevertheless, the performance evaluation under dataset shift highlights that only the SF method can cope and exhibit, to some extent, transfer learning capabilities between different domains/environments. Finally, the memory requirements and the combined processing and inference time analysis of each learning representation demonstrate that the enhanced detection performance of the LF-ASH scheme, as compared to the other methods, comes at a cost due to the higher volume of processed information. 
\subsection{Performance Evaluation on Real-Life Object Detector}

{In addition to Faster-RCNN and FCOS, we expand the evaluation of the introspection mechanism to include YOLOv8, a prevalent one-stage object detector with a wide use particularly in ADS domain. For a comprehensive analysis, we selected the highest-performing mechanism among the SOTA, i.e., the SF, and also incorporated the raw, unprocessed activations (LFR). This dual approach allows us to not only benchmark the proposed mechanism against established methods in the literature but also to understand the implications of activation shaping and its absence on the mechanism's performance. Similar to the experiments with base detectors Faster-RCNN and FCOS, a hyperparameter tuning with same search space has been employed for this experiment, and the best parameters for each case are presented in Table~\ref{tab:yoloparams}. }
\begin{table}[t]
\begin{center}
\caption{Best parameters obtained from hyperparameter tuning for each configuration of dataset, and selected introspection model for YOLOv8 detector.}
\label{tab:yoloparams}
\begin{tabular}{ccccc}
\textbf{Dataset} &\textbf{Method} & \makecell{\textbf{Batch}\\\textbf{Size}} & \makecell{\textbf{Focal} \textbf{Loss}\\\textbf{Gamma}} & \textbf{LR} \\ \toprule
\multirow{3}{*}{BDD} & {LF-ASH}   & {32}   & {0} & {0.001} \\      
      &         {LFR}   & {64}   & {2} & {0.001} \\
      &          {SF}   & {16}   & {0} & {0.001} \\    \midrule 
\multirow{3}{*}{KITTI}&   {LF-ASH}   & {32}   & {5} & {0.001} \\      
              &  {LFR}   & {16}   & {5} & {0.001} \\
              &  {SF}   & {16}   & {5} & {0.001} \\    \bottomrule
\end{tabular}
\end{center}
\end{table}

Parameters in Table~\ref{tab:yoloparams} show that similar to the previous sections due to the stronger imbalance in KITTI dataset the gamma value is higher in KITTI scenario, but majority of other parameters are more or less similar, batch size varying between 16-64 for BDD, and 16-32 in KITTI, and learning rate being 0.001 for all. On the other hand, it is important to note that in the YOLOv8 experiments, the Adam optimiser was identified as the optimal choice for the BDD dataset across all learning representations unlike the other presented parameters in previous section. The remaining cases utilised a stochastic gradient descent optimiser as the previous section denoted.
\begin{table}[t]
\begin{center}
\caption{Performance evaluation of each selected introspection method on selected datasets for YOLOv8 detector.}
\label{tab:yoloperf}
\begin{tabular}{ccccc}
 \makecell{\textbf{Dataset}}  & \textbf{Method} & \textbf{AUROC} & \textbf{F1} & \textbf{FNR} \\ \toprule
 \multirow{3}{*}{BDD}& LF-ASH &{0.8176}&{0.6319}&{0.3261}\\ 		
                        & LFR    &{0.8142}&{0.6667}&{0.2322}\\
                        & SF     &{0.7685}&{0.6119}&{0.3208}\\ \midrule
\multirow{3}{*}{KITTI}&  LF-ASH &{0.9200}&{0.7500}&{0.0827}\\
                        &                          LFR    &{0.8563}&{0.5627}&{0.3062}\\
                        &                         SF     &{0.9006}&{0.6326}&{0.2068}\\\bottomrule
\end{tabular}
\end{center}
\end{table}

{
Table~\ref{tab:yoloperf} demonstrates distinct performance metrics across two datasets. In the BDD dataset, LF-ASH leads with the highest AUROC (0.8176), while LFR exhibits the best FNR (0.232) and a competitive F1 score. SF has a modest AUROC but the best F1 score. In the KITTI dataset, LF-ASH outperforms in all metrics, with SF and LFR following in descending order of AUROC, F1, and FNR. This analysis suggests that while LF-ASH generally shows superior performance, especially in the KITTI dataset, trade-offs between different metrics like FNR and F1 score are evident} 

{For the cross-dataset performance experiments with YOLOv8 detector also showed that the transferability of the information across datasets is limited, and only the SF mechanism maintains to some extent its performance without losing the balance between safe and non safe samples. However, when the AUROC values compared with the previous samples, we see worse capability for SF, highlighting that the information transfer capability of SF is not as good as in the case of other detectors. The other two methods show similar patterns with small changes in the metrics, but still classifying most samples to a single class as in the previous results. Lastly, the transferring information from larger and more complex dataset, BDD, to a smaller one, KITTI, is better despite the overall low performance.}
\begin{table}[t]
\begin{center}
\caption{Performance evaluation of each selected introspection method on selected datasets for YOLOv8 detector.}
\label{tab:yoloshift}
\begin{tabular}{cccccc}
\makecell{\textbf{Train}\\\textbf{Set}}& \makecell{\textbf{Test}\\\textbf{Set}} & \textbf{Method} & \textbf{AUROC} & \textbf{F1} & \textbf{FNR} \\ \toprule
\multirow{3}{*}{BDD}   & \multirow{3}{*}{KITTI} & LF-ASH &{0.5352}&{0.4605}&{0.4791}\\
&  & LFR    &{0.5931}&{0.4795}&{0.0000}\\
&  & SF     &{0.6727}&{0.5670}&{0.3172}\\  \midrule
  \multirow{3}{*}{KITTI}   & \multirow{3}{*}{BDD} & {LF-ASH} &{0.5162}&{0.5279}&{0.1279}\\
   &  & {LFR}    &{0.5029}&{0.4530}&{0.0000}        \\
   &  & {SF}     &{0.5253} &{0.3416}&{0.6206}\\
\bottomrule
\end{tabular}
\end{center}
\end{table}

{Overall, the LF-ASH mechanism demonstrates either the best or highly competitive results across both datasets, despite some variations in specific cases. However, its performance in the BDD dataset falls short as compared to the outcomes attained using the Faster-RCNN and FCOS detectors. This discrepancy might stem from the preliminary analysis conducted on FCOS for mode and percentile selection, i.e., recall that our proposed mechanism is using the mode P, pruning the activations based on the percentile value only, with 75th percentile for BDD and 90th percentile for KITTI dataset.  Similar performance trends are observable in Faster-RCNN, attributable to its shared backbone with FCOS, i.e., ResNet50, a feature not present in YOLOv8. Consequently, additional research is warranted to explore if some better mode-percentile pairing exists for YOLOv8, a task beyond the current study's scope. Additionally, we did not present a computational requirement analysis, as the primary change in YOLO pertains to the size of the activation maps. This alteration is expected to result in a reduction across the computational performance metrics, but the relative comparison between different learning representations, as discussed in the previous section, is anticipated to remain consistent.}

\subsection{Qualitative Performance Evaluation of the Proposed Introspection Mechanism}
\begin{figure*}[t]
    \centering
    \includegraphics[width=\linewidth]{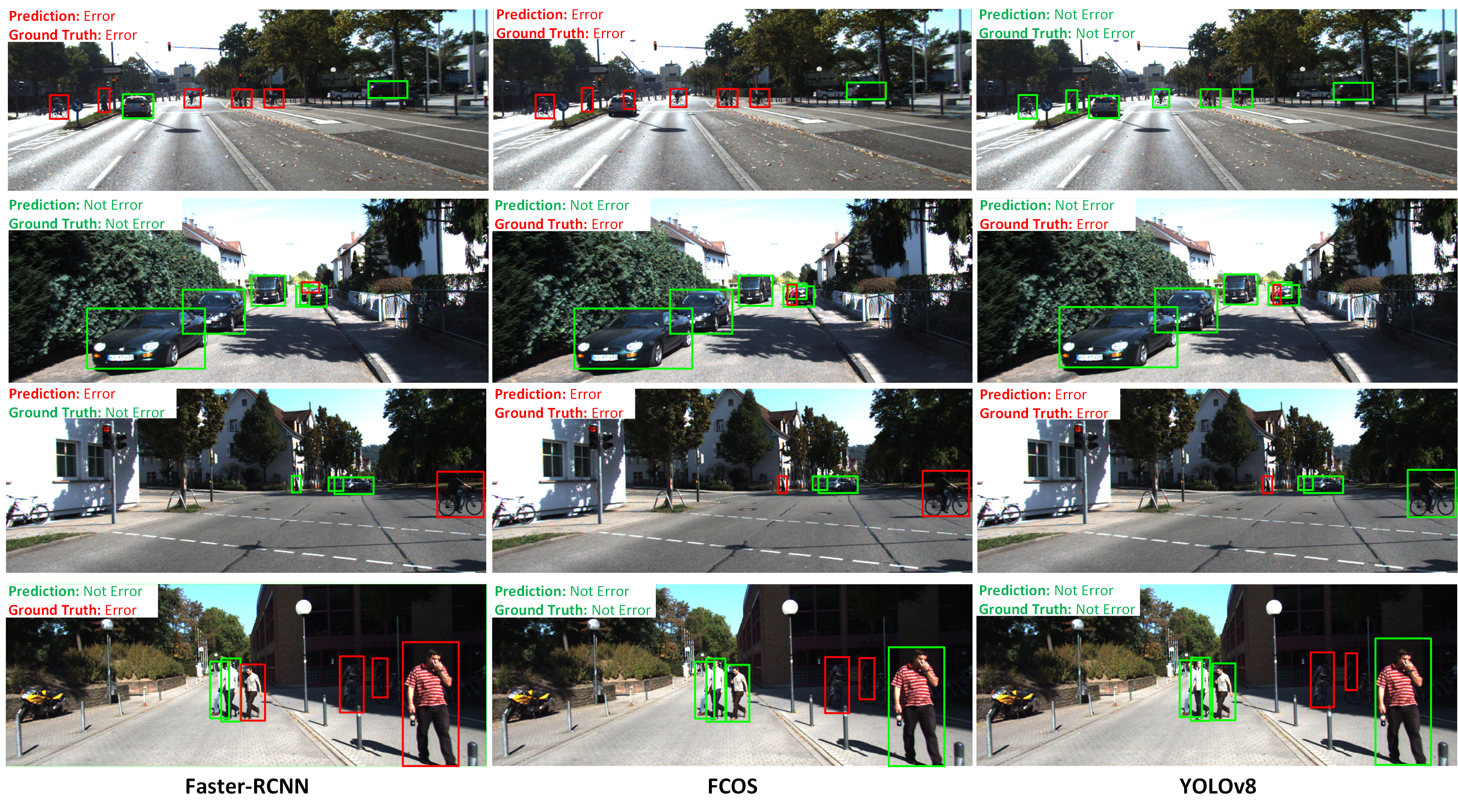}
    \caption{{This figure presents the introspection mechanism's performance variations across three distinct object detection models—Faster R-CNN (left column), FCOS (middle), and YOLOv8 (right) — each represented by a separate column, applied to the same urban street scenes taken from KITTI dataset. While the occurrence of missed objects and the resultant error labels may vary between detectors, a degree of similarity is observed among one-stage detectors like FCOS and YOLOv8. Annotations highlight the correct and incorrect predictions of the object detection functionality: red boxes indicate false positives or false negatives, indicating detection errors, whereas green boxes denote true positives. A sample is categorised as 'non-error' in the error dataset if its mean Average Precision (mAP) surpasses the set threshold, and 'error' if it does not. The introspection model's predictions, along with the ground truth, are detailed in the upper left corner of each image, enhancing the understanding of each model's detection capabilities and the introspection's result.}}
    \label{fig:qual}
\end{figure*}

{Apart from the quantitative results presented in the previous sections, qualitative analysis per frame can also be used to showcase the strengths and weaknesses of the introspection mechanisms.  Figure~\ref{fig:qual} presents four such cases with frames taken from the KITTI dataset. At the top left corner of each frame, the introspection mechanism's prediction and the ground truth label are indicated. A matching of the two indicates that the introspection model has correctly learned to associate the mAP with the extracted features of the object detector. Recall that the ground truth label is classified as erroneous when the calculated mAP is less than the threshold and as non-erroneous otherwise.  True positives are marked in green bounding boxes, while false negatives and false negatives are marked in red bounding boxes.}

{In the first scene, Faster-RCNN successfully detects all cars but misses pedestrians and cyclists, resulting in a mAP lower than the  threshold, thus indicating an evident error. This effectively demonstrates that numerous undetected objects correspond to a low mAP, simplifying error detection for the proposed mechanism. The introspection model engaged with  Faster-RCNN is capable of recognising its underperformance and classify the frame as erroneous. A similar pattern of missed detections, including vehicles, is noted with the FCOS detector, while YOLOv8  exhibits no misses, underlying the variability in detector performance in the same scenarios.}

{In the second scene, one-stage detectors (YOLOv8 and FCOS)  miss the person, dropping the average precision for that class and thus, the  mAP for that frame. Conversely, Faster-RCNN does not replicate this error because it only misses a distant vehicle. We see that identifying the effect of missing pedestrian on mAP is challenging to capture through the activation maps due to the high number of correct detections in that frame, the individual's distance from the ego vehicle, and their proximity to other detected objects. This is the reason that with activations from YOLOv8 and FCOS  the introspection mechanism fails to predict the error. }

{In the third scene, the error label depends on the pedestrian and the vehicles (including cyclist). In such instances, our model associated with Faster-RCNN cannot make a correct prediction when the missed object(s) are large. In contrast, for FCOS and YOLOv8 the introspection model is capable of capturing the errors of the detector caused by small objects sufficiently isolated from other objects, allowing the mechanism to recognise discernible patterns. }

{The last scene includes pedestrians and a cyclist, with one pedestrian and the cyclist being hard to discern due to poor illumination. The Faster-RCNN detector fails to detect one of the pedestrians on the street along with the cyclist and the pedestrian in dimly lit areas. Despite these misses, the introspection mechanism categorizes the scene as "not-error," suggesting that misdetections in darker regions have a minimal impact on our mechanism's assessment. Conversely, when objects in well-lit areas are successfully detected, the mechanism accurately recognizes these as "not error" with the other two detectors, highlighting its sensitivity to illumination levels in object detection.}

{In conclusion, the qualitative results reveal that the introspection mechanism is influenced by diverse factors, including scene context and detector capabilities. Key challenges in 2D object detection, like low illumination and small object detection, can significantly affect introspection. Our analysis also show the need for thorough error criterion investigation, as mAP values can drastically change with just one misdetection, altering scene labels.  Furthermore, as seen in the first row of Figure~\ref{fig:qual} there are some objects far away neither highlighted as true positive or false negative. This is due to the labelling of the datasets, which sometimes might be erroneous or partial \cite{valentinaerror}. It is also possible that such cases may confuse the introspection mechanisms as the activations may indicate the presence of an object in those areas.}

\subsection{Effect of Performance Metric Threshold on Real-Life Object Detectors}
{The criterion for generating the error datasets have been so far defined using a threshold value of 0.5 for the mAP metric, in line with the approach used by Rahman et al. in~\cite{rahmanper}. This value was selected to introduce some level of tolerance for errors associated with small or distant objects. Next, we examine how adjusting this threshold influences the error detection performance of YOLOv8 along with the proposed mechanism for introspection. We investigate threshold values ranging from 0.4 to 0.7 in increments of 0.1. We confined our analysis to this range because thresholds larger than 0.7 or below 0.4 significantly increase the existing data imbalance, to a level where learning-based mechanisms struggle to perform effectively. Additionally, it is important to highlight that the same hyperparameter tuning mentioned in Section~\ref{sec:detperf} has been applied for this process.}
\begin{table}[h]
\centering
\caption{Performance Evaluation of the Proposed Introspection Mechanism using YOLOv8 Detector on KITTI and BDD Datasets.}\label{tab:sens}
\begin{tabular}{ccccc}
{\textbf{Dataset}} & \textbf{Threshold} & \textbf{AUROC} & \textbf{F1 Score} & \textbf{FNR} \\\toprule
\multirow{4}{*}{KITTI} & 0.4 & 0.9499 & 0.6792 & 0.0654 \\
                       & 0.5 & 0.9200 & 0.7500     & 0.0827 \\
                       & 0.6 & 0.8412 & 0.5640 & 0.3649 \\
                       & 0.7 & 0.6384 & 0.5024 & 0.4433 \\ \midrule
\multirow{4}{*}{BDD}   & 0.4 & 0.8645 & 0.5515 & 0.3888 \\
                       & 0.5 & 0.8176 & 0.6319 & 0.3261 \\
                       & 0.6 & 0.7975  & 0.6658 & 0.3856 \\
                       & 0.7 & 0.7502 & 0.6820 & 0.2841\\ \bottomrule
\end{tabular}
\end{table}

{Table~\ref{tab:sens} shows the performance evaluation for varying mAP threshold values (0.4, 0.5, 0.6, 0.7). Recall that mAP scores lower than the selected threshold are used to classify the frame as erroneous. In the KITTI dataset, the optimal F1 Score of 0.75 is observed at a threshold of 0.5. However, this is accompanied by an elevated False Negative Rate (FNR) of 0.0827, in contrast to a lower FNR of 0.0654 at the 0.4 threshold. Notably, the AUROC  diminishes substantially with increasing thresholds, descending to 0.6384 at a threshold of 0.7. Conversely, in the BDD dataset, the F1 Score escalates with higher thresholds, reaching a peak of 0.6820 at 0.7. Despite this, there is a consistent decline in AUROC values. The minimum FNR for BDD, recorded at 0.2841, is also at the 0.7 threshold, indicating a more balanced trade-off between sensitivity and specificity at this threshold for BDD.}

{The primary rationale for these variations is attributed to the shifting balance of imbalance with changing thresholds and the distinct characteristics of errors at each threshold. An increased threshold induces the model to concentrate on intricate patterns, facilitating the identification of nuanced errors, which may not be directly related to safety, such as distant objects. Conversely, excessively lowering the threshold poses challenges in learning patterns, as a substantial number of samples will be attributed as non-errors, leading the model to oversee some critical errors. A comprehensive analysis suggests that a 0.5 threshold represents an optimal compromise when considering all metrics collectively. Nevertheless, it is imperative to highlight the necessity for further exploration alongside augmented research into error criteria (beyond the mAP) to enhance the robustness and efficacy of introspection in object detection models.}

\section{Discussion}\label{sec:disc}
The experiments presented in the previous section provided insights into the performance, computational and memory requirements of several learning representations for error detection in ADS. Different representations exhibited distinct strengths and weaknesses. 
Our results suggest that leveraging learned features from activation shaping (LF-ASH) to identify erroneous detections in a frame has the potential to enhance the safety of ADS perception. In particular, simple pruning of the activation maps appears to yield the best error detection performance. Using statistical functions (SF) to extract features from the raw activations also showed promising error detection performance, while the computational and memory requirements were kept under that of LF-ASH. These findings have practical implications for the design of error detection systems in ADS, which need to balance the trade-off between perception accuracy and resource efficiency. 
We also observed that using LF-ASH may not adapt well to dataset shifts compared to SF. This implies a need to further simplify the learned features, but a certain level of complexity is also necessary to maintain consistency within datasets and provide reliable error detection. 


It is also important to acknowledge certain limitations in this study. Firstly, our analysis only covered a limited set of learning representations, and modifications were made to some baselines for a fair comparison. {Secondly, despite the investigation presented in this paper, the selected mAP threshold for indicating faults should be further investigated to understand its effect on different learning representations. Similarly, it is crucial to note that the mAP metric-based error criteria might not recognise certain scenes as errors where an object detector's prediction could lead to high-risk situations. This stems from the nature of mAP, which calculates the average of the Average Precision (AP) across different classes. This averaging process means that mAP generalises performance by balancing the detection accuracy among all classes. Additionally, it's important to note that the AP for each class is influenced by the number of samples in that class. Thus, in cases where certain classes have fewer samples, their impact on the overall mAP might be less pronounced compared to classes with more samples. mAP focuses on the ability to detect multiple objects within their classes, which may overlook scenarios where a single object, such as a car, is missed in a critical position within the scene. For example, the sample in the first column, third row of Figure~\ref{fig:qual} is labelled as non-erroneous, yet overlooking the cyclist could potentially lead to a crash, creating a high-risk situation. Therefore, although it is a common metric to assess performance, and used in introspection, there is a need for further research into more context- and environment-specific error criteria.} Alternatively, it may be possible to structure the introspection mechanism as a regression or multi-class classification problem to differentiate errors based on the error severity. Thirdly, our evaluation was performed on a specific set of datasets. Further studies are therefore necessary to assess the generalisation of the findings to other datasets with larger sample sizes. {Similarly, while the chosen object detectors exemplify both one- and two-stage detection methods, additional research is required on other widely-used detectors, and recently introduced transformer-based detectors that have not yet seen widespread use in ADS, such as DETR \cite{detr}.} This is necessary to ensure the generalisability of the proposed introspection mechanism. Finally, the computational and memory requirements were analysed on a single device. More experimentation is needed to reveal how these requirements would scale up to larger systems with more restricted resources. Future research should aim to address the above limitations and further explore the potential of utilising neural network activations for ADS perception safety.

\section{Summary \& Conclusion}\label{sec:conclusion}

This paper introduced a novel introspection mechanism that utilises (shaped) activations patterns as a learning representation to identify frame-level patterns causing 2D object detector failures in ADS. To compare its performance with SOTA metric-based introspection methods, we devised a unified four-stage framework including object detector training, error dataset generation, introspection training, and testing. We tested this framework both with one- and two-stage object detectors and evaluated all introspection methods with errors extracted from two well-known driving datasets, KITTI and BDD.
Our results show that utilising shaped neural network activations with performance metric-based introspection is promising for detecting perception system faults in ADS. We have tried several shaping methods concluding that simply pruning the neural network activation maps yields the best error detection performance. We also highlight that extracting statistical features before learning erroneous patterns provides competitive overall performance and adaptability at the cost of missed detection errors (false negatives), which are however critical from a safety point of view. There are admittedly further challenges to tackle, mainly striking a balance between error detection performance under the same dataset and in dataset-shift cases, as well as reducing the computational requirements of the selected introspection method. Future research will investigate the above-mentioned trade-off to achieve more robust, reliable and less resource-intensive error detection for ADS perception.

\bibliographystyle{IEEEtran}
\bibliography{main.bib}

\begin{thebibliography}{10}
\providecommand{\url}[1]{#1}
\csname url@samestyle\endcsname
\providecommand{\newblock}{\relax}
\providecommand{\bibinfo}[2]{#2}
\providecommand{\BIBentrySTDinterwordspacing}{\spaceskip=0pt\relax}
\providecommand{\BIBentryALTinterwordstretchfactor}{4}
\providecommand{\BIBentryALTinterwordspacing}{\spaceskip=\fontdimen2\font plus
\BIBentryALTinterwordstretchfactor\fontdimen3\font minus \fontdimen4\font\relax}
\providecommand{\BIBforeignlanguage}[2]{{%
\expandafter\ifx\csname l@#1\endcsname\relax
\typeout{** WARNING: IEEEtran.bst: No hyphenation pattern has been}%
\typeout{** loaded for the language `#1'. Using the pattern for}%
\typeout{** the default language instead.}%
\else
\language=\csname l@#1\endcsname
\fi
#2}}
\providecommand{\BIBdecl}{\relax}
\BIBdecl

\bibitem{nhtsa:2018t}
{U.S National Transportation Safety Board}, ``Collision between a sport utility vehicle operating with partial driving automation and a crash attenuator,'' \url{www.ntsb.gov/investigations/AccidentReports/Reports/HAR2001.pdf}, Feb 2020.

\bibitem{hakansurvey}
H.~Y. Yatbaz, M.~Dianati, and R.~Woodman, ``Introspection of dnn-based perception functions in automated driving systems: State-of-the-art and open research challenges,'' \emph{IEEE Transactions on Intelligent Transportation Systems}, 2023.

\bibitem{sae}
\BIBentryALTinterwordspacing
``Taxonomy and definitions for terms related to driving automation systems for on-road motor vehicles.'' [Online]. Available: \url{https://doi.org/10.4271/j3016\_202104}
\BIBentrySTDinterwordspacing

\bibitem{koopman2016challenges}
P.~Koopman and M.~Wagner, ``Challenges in autonomous vehicle testing and validation,'' \emph{SAE International Journal of Transportation Safety}, vol.~4, no.~1, pp. 15--24, 2016.

\bibitem{detectionsurvey}
S.~S.~A. Zaidi, M.~S. Ansari, A.~Aslam, N.~Kanwal, M.~Asghar, and B.~Lee, ``A survey of modern deep learning based object detection models,'' \emph{Digital Signal Processing}, vol. 126, p. 103514, 2022.

\bibitem{valentinanoise}
B.~Li, P.~H. Chan, G.~Baris, M.~D. Higgins, and V.~Donzella, ``Analysis of automotive camera sensor noise factors and impact on object detection,'' \emph{IEEE Sensors Journal}, vol.~22, no.~22, pp. 22\,210--22\,219, 2022.

\bibitem{dropoutod}
D.~Miller, L.~Nicholson, F.~Dayoub, and N.~Sunderhauf, ``\BIBforeignlanguage{English}{Dropout sampling for robust object detection in open-set conditions},'' in \emph{\BIBforeignlanguage{English}{Proceedings - IEEE International Conference on Robotics and Automation}}, 2018, pp. 3243--3249.

\bibitem{rahmanper}
Q.~M. Rahman, N.~Sünderhauf, and F.~Dayoub, ``Per-frame map prediction for continuous performance monitoring of object detection during deployment,'' in \emph{2021 IEEE Winter Conference on Applications of Computer Vision Workshops (WACVW)}, 2021, pp. 152--160.

\bibitem{fnyang}
\BIBentryALTinterwordspacing
Q.~Yang, H.~Chen, Z.~Chen, and J.~Su, ``Introspective false negative prediction for black-box object detectors in autonomous driving,'' \emph{Sensors}, vol.~21, no.~8, 2021. [Online]. Available: \url{https://www.mdpi.com/1424-8220/21/8/2819}
\BIBentrySTDinterwordspacing

\bibitem{posthoc2022}
X.~Zhang, S.~Oymak, and J.~Chen, ``Post-hoc models for performance estimation of machine learning inference,'' \emph{arXiv preprint arXiv:2110.02459}, 2021.

\bibitem{ash}
\BIBentryALTinterwordspacing
A.~Djurisic, N.~Bozanic, A.~Ashok, and R.~Liu, ``Extremely simple activation shaping for out-of-distribution detection,'' 2022. [Online]. Available: \url{https://arxiv.org/abs/2209.09858}
\BIBentrySTDinterwordspacing

\bibitem{hakaniccv}
H.~Y. Yatbaz, M.~Dianati, K.~Koufos, and R.~Woodman, ``Introspection of 2d object detection using processed neural activation patterns in automated driving systems,'' in \emph{Proceedings of the IEEE/CVF International Conference on Computer Vision (ICCV) Workshops}, October 2023, pp. 4047--4054.

\bibitem{fcos}
Z.~Tian, C.~Shen, H.~Chen, and T.~He, ``Fcos: Fully convolutional one-stage object detection,'' in \emph{Proceedings of the IEEE/CVF international conference on computer vision}, 2019, pp. 9627--9636.

\bibitem{yolo}
\BIBentryALTinterwordspacing
G.~Jocher, A.~Chaurasia, and J.~Qiu, ``{Ultralytics YOLO},'' Jan. 2023. [Online]. Available: \url{https://github.com/ultralytics/ultralytics}
\BIBentrySTDinterwordspacing

\bibitem{fasterrcnn}
S.~Ren, K.~He, R.~Girshick, and J.~Sun, ``Faster r-cnn: Towards real-time object detection with region proposal networks,'' \emph{IEEE Transactions on Pattern Analysis and Machine Intelligence}, vol.~39, no.~6, pp. 1137--1149, 2017.

\bibitem{kitti}
A.~Geiger, P.~Lenz, C.~Stiller, and R.~Urtasun, ``Vision meets robotics: The kitti dataset,'' \emph{The International Journal of Robotics Research}, vol.~32, no.~11, pp. 1231--1237, 2013.

\bibitem{bdd100k}
F.~Yu, H.~Chen, X.~Wang, W.~Xian, Y.~Chen, F.~Liu, V.~Madhavan, and T.~Darrell, ``Bdd100k: A diverse driving dataset for heterogeneous multitask learning,'' in \emph{The IEEE Conference on Computer Vision and Pattern Recognition (CVPR)}, June 2020.

\bibitem{bayesod}
A.~Harakeh, M.~Smart, and S.~L. Waslander, ``\BIBforeignlanguage{English}{Bayesod: A bayesian approach for uncertainty estimation in deep object detectors},'' in \emph{\BIBforeignlanguage{English}{Proceedings - IEEE International Conference on Robotics and Automation}}, 2020.

\bibitem{mcdropout}
Y.~Gal and Z.~Ghahramani, ``Dropout as a bayesian approximation: Representing model uncertainty in deep learning,'' in \emph{International Conference on Machine Learning}.\hskip 1em plus 0.5em minus 0.4em\relax PMLR, 2016, pp. 1050--1059.

\bibitem{ssd}
W.~Liu, D.~Anguelov, D.~Erhan, C.~Szegedy, S.~Reed, C.-Y. Fu, and A.~C. Berg, ``Ssd: Single shot multibox detector,'' in \emph{Computer Vision -- ECCV 2016}, B.~Leibe, J.~Matas, N.~Sebe, and M.~Welling, Eds.\hskip 1em plus 0.5em minus 0.4em\relax Cham: Springer International Publishing, 2016, pp. 21--37.

\bibitem{gmmdet}
D.~Miller, N.~S{\"u}nderhauf, M.~Milford, and F.~Dayoub, ``Uncertainty for identifying open-set errors in visual object detection,'' \emph{IEEE Robotics and Automation Letters}, vol.~7, no.~1, pp. 215--222, 2021.

\bibitem{pixelinv}
Y.~Wang and D.~Wijesekera, ``Pixel invisibility: Detecting objects invisible in color images,'' \emph{arXiv preprint arXiv:2006.08383}, 2020.

\bibitem{du2022unknown}
X.~Du, X.~Wang, G.~Gozum, and Y.~Li, ``Unknown-aware object detection: Learning what you don't know from videos in the wild,'' in \emph{Proceedings of the IEEE/CVF Conference on Computer Vision and Pattern Recognition}, 2022, pp. 13\,678--13\,688.

\bibitem{9922026}
C.~Huang, V.~D. Nguyen, V.~Abdelzad, C.~G. Mannes, L.~Rowe, B.~Therien, R.~Salay, and K.~Czarnecki, ``Out-of-distribution detection for lidar-based 3d object detection,'' in \emph{2022 IEEE 25th International Conference on Intelligent Transportation Systems (ITSC)}, 2022, pp. 4265--4271.

\bibitem{9665821}
J.~Cen, P.~Yun, J.~Cai, M.~Y. Wang, and M.~Liu, ``Open-set 3d object detection,'' in \emph{2021 International Conference on 3D Vision (3DV)}, 2021, pp. 869--878.

\bibitem{rahmancascade}
Q.~M. Rahman, N.~S{\"u}nderhauf, and F.~Dayoub, ``Online monitoring of object detection performance post-deployment,'' \emph{arXiv preprint arXiv:2011.07750}, 2020.

\bibitem{failingtolearn}
M.~S. Ramanagopal, C.~Anderson, R.~Vasudevan, and M.~Johnson-Roberson, ``Failing to learn: Autonomously identifying perception failures for self-driving cars,'' \emph{IEEE Robotics and Automation Letters}, vol.~3, no.~4, pp. 3860--3867, 2018.

\bibitem{kdiagnose}
P.~Antonante, D.~I. Spivak, and L.~Carlone, ``Monitoring and diagnosability of perception systems,'' in \emph{2021 IEEE/RSJ International Conference on Intelligent Robots and Systems (IROS)}.\hskip 1em plus 0.5em minus 0.4em\relax IEEE, 2021, pp. 168--175.

\bibitem{antonante2022monitoring}
P.~Antonante, H.~Nilsen, and L.~Carlone, ``Monitoring of perception systems: Deterministic, probabilistic, and learning-based fault detection and identification,'' \emph{Artificial Intelligence}, p. 103998, 2023.

\bibitem{Hawke2016}
J.~Hawke, C.~Gur{\u{a}}u, C.~H. Tong, and I.~Posner, ``Wrong today, right tomorrow: Experience-based classification for robot perception,'' in \emph{Field and Service Robotics}.\hskip 1em plus 0.5em minus 0.4em\relax Springer, 2016, pp. 173--186.

\bibitem{fitforpurpose}
C.~Gur{\u{a}}u, C.~H. Tong, and I.~Posner, ``Fit for purpose? predicting perception performance based on past experience,'' in \emph{International Symposium on Experimental Robotics}.\hskip 1em plus 0.5em minus 0.4em\relax Springer, 2016, pp. 454--464.

\bibitem{learnfromexp}
C.~Gur{\u{a}}u, D.~Rao, C.~H. Tong, and I.~Posner, ``Learn from experience: Probabilistic prediction of perception performance to avoid failure,'' \emph{The International Journal of Robotics Research}, vol.~37, no.~9, pp. 981--995, 2018.

\bibitem{coco}
T.-Y. Lin, M.~Maire, S.~Belongie, J.~Hays, P.~Perona, D.~Ramanan, P.~Doll{\'a}r, and C.~L. Zitnick, ``Microsoft coco: Common objects in context,'' in \emph{Computer Vision--ECCV 2014: 13th European Conference, Zurich, Switzerland, September 6-12, 2014, Proceedings, Part V 13}.\hskip 1em plus 0.5em minus 0.4em\relax Springer, 2014, pp. 740--755.

\bibitem{pascalvoc}
M.~Everingham, L.~Van~Gool, C.~K.~I. Williams, J.~Winn, and A.~Zisserman, ``The pascal visual object classes (voc) challenge,'' \emph{International Journal of Computer Vision}, vol.~88, no.~2, pp. 303--338, Jun. 2010.

\bibitem{autoware}
\BIBentryALTinterwordspacing
{Autoware Foundation}, ``Autoware.universe.'' [Online]. Available: \url{https://github.com/autowarefoundation/autoware.universe/tree/v0.8.0}
\BIBentrySTDinterwordspacing

\bibitem{classweight}
G.~King and L.~Zeng, ``Logistic regression in rare events data,'' \emph{Political analysis}, vol.~9, no.~2, pp. 137--163, 2001.

\bibitem{energy}
W.~Liu, X.~Wang, J.~Owens, and Y.~Li, ``Energy-based out-of-distribution detection,'' \emph{Advances in neural information processing systems}, vol.~33, pp. 21\,464--21\,475, 2020.

\bibitem{focal}
T.-Y. Lin, P.~Goyal, R.~Girshick, K.~He, and P.~Dollár, ``Focal loss for dense object detection,'' in \emph{2017 IEEE International Conference on Computer Vision (ICCV)}, 2017, pp. 2999--3007.

\bibitem{valentinaerror}
B.~Li, G.~Baris, P.~H. Chan, A.~Rahman, and V.~Donzella, ``Testing ground-truth errors in an automotive dataset for a dnn-based object detector,'' in \emph{2022 International Conference on Electrical, Computer, Communications and Mechatronics Engineering (ICECCME)}, 2022, pp. 1--6.

\bibitem{detr}
N.~Carion, F.~Massa, G.~Synnaeve, N.~Usunier, A.~Kirillov, and S.~Zagoruyko, ``End-to-end object detection with transformers,'' in \emph{European conference on computer vision}.\hskip 1em plus 0.5em minus 0.4em\relax Springer, 2020, pp. 213--229.

\end{thebibliography}

\end{document}